\documentclass{article} % For LaTeX2e
\usepackage[table]{xcolor}
\usepackage{iclr2025_conference,times}
\usepackage[whole]{bxcjkjatype}

\usepackage{amsmath,amsfonts,bm}

\def\eqref#1{equation~\ref{#1}}
\def\1{\bm{1}}

\DeclareMathAlphabet{\mathsfit}{\encodingdefault}{\sfdefault}{m}{sl}
\SetMathAlphabet{\mathsfit}{bold}{\encodingdefault}{\sfdefault}{bx}{n}

\usepackage{hyperref}
\usepackage{url}
\usepackage{pdfpages}
\usepackage{graphicx}
\usepackage{float}
\usepackage{booktabs}
\usepackage{multirow}
\usepackage{makecell}

\usepackage{adjustbox}
\usepackage{threeparttable}
\definecolor{light-gray}{gray}{0.9}
\usepackage{algorithm}
\definecolor{verylightgray}{rgb}{0.93, 0.93, 0.93}

\newcommand{\methodname}{Drop-Upcycling}
\newcommand{\NUname}{na\"{i}ve Upcycling} 
\newcommand{\RNUname}{Random Noise Upcycling} 
\newcommand{\diff}[1]{#1}

\usepackage{xspace}
\newcommand{\huggingface}{\raisebox{-1.5pt}{\includegraphics[height=1.05em]{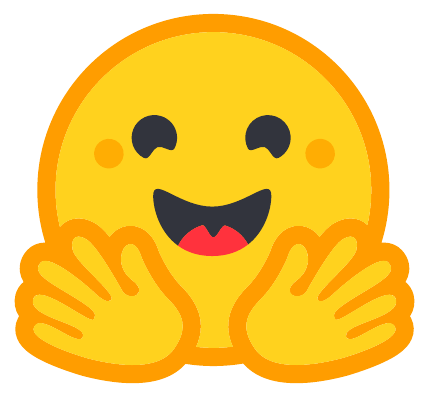}}\xspace}
\newcommand{\github}{\raisebox{-1.5pt}{\includegraphics[height=1.05em]{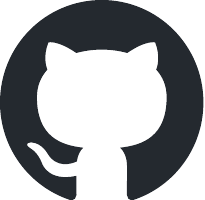}}\xspace}
\newcommand{\wandb}{\raisebox{-1.5pt}{\includegraphics[height=1.05em]{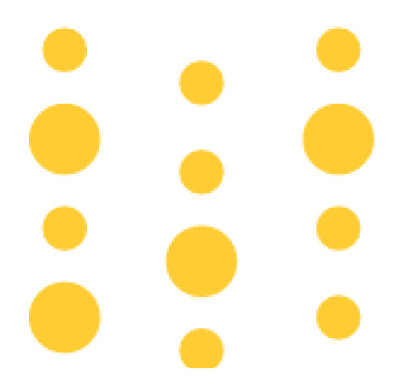}}\xspace}
\newcommand{\gitlab}{\raisebox{-1.5pt}{\includegraphics[height=1.05em]{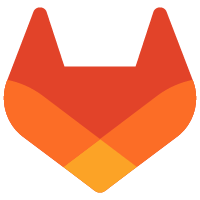}}\xspace}

\title{
\methodname{}: Training Sparse Mixture of Experts with Partial Re-initialization
}

\author{
Taishi Nakamura$^{1,2,3}$, 
Takuya Akiba$^{2}$, 
Kazuki Fujii$^{1}$,
Yusuke Oda$^{3}$, \\
\,\,\textbf{Rio Yokota}$^{1,3}$, 
\textbf{Jun Suzuki}$^{4,5,3}$
\\
$^1$Institute of Science Tokyo,
$^2$Sakana AI,
$^3$NII LLMC,
$^4$Tohoku University,
$^5$RIKEN\\
\texttt{taishi@rio.scrc.iir.isct.ac.jp},
\texttt{jun.suzuki@tohoku.ac.jp}
}

\iclrfinalcopy % Uncomment for camera-ready version, but NOT for submission.
\begin{document}

\maketitle
\begin{abstract}
The Mixture of Experts (MoE) architecture reduces the training and inference cost significantly compared to a dense model of equivalent capacity. Upcycling is an approach that initializes and trains an MoE model using a pre-trained dense model. While upcycling leads to initial performance gains, the training progresses slower than when trained from scratch, leading to suboptimal performance in the long term. We propose \emph{\methodname{}} -- a method that effectively addresses this problem. \methodname{} combines two seemingly contradictory approaches: utilizing the knowledge of pre-trained dense models while statistically re-initializing some parts of the weights. This approach strategically promotes expert specialization, significantly enhancing the MoE model's efficiency in knowledge acquisition. 
Extensive large-scale experiments demonstrate that \methodname{} significantly outperforms previous MoE construction methods in the long term, specifically when training on hundreds of billions of tokens or more.
As a result, our MoE model with 5.9B active parameters achieves comparable performance to a 13B dense model in the same model family, while requiring approximately 1/4 of the training FLOPs.
% This research offers new insights for efficient LLM development and understanding of MoE models.
All experimental resources, including source code, training data, model checkpoints and logs, are publicly available to promote reproducibility and future research on MoE.

\begin{center}
\begin{tabular}{rcl} % This defines 3 columns: right-aligned, center-aligned, left-aligned
\huggingface & \textbf{Weights} & \href{https://huggingface.co/collections/llm-jp/drop-upcycling-674dc5be7bbb45e12a476b80}{\path{huggingface.co/collections/llm-jp/}}\\
& & \href{https://huggingface.co/collections/llm-jp/drop-upcycling-674dc5be7bbb45e12a476b80}{\path{drop-upcycling-674dc5be7bbb45e12a476b80}}\\[0.2em]
\gitlab & \textbf{Data} & \href{https://gitlab.llm-jp.nii.ac.jp/datasets/llm-jp-corpus-v3}{\path{gitlab.llm-jp.nii.ac.jp/}}\\
& & \href{https://gitlab.llm-jp.nii.ac.jp/datasets/llm-jp-corpus-v3}{\path{datasets/llm-jp-corpus-v3}}\\[0.2em]
\github & \textbf{Code} & \href{https://github.com/Taishi-N324/Drop-Upcycling}{\path{github.com/Taishi-N324/Drop-Upcycling}}\\[0.2em]
\wandb & \textbf{Logs} & \href{https://wandb.ai/taishi-nakamura/Drop-Upcycling}{\path{wandb.ai/taishi-nakamura/Drop-Upcycling}}
\end{tabular}
\end{center}
\end{abstract}

\section{Introduction}

%
% LLM's training and inference are too expensive
%
Large-scale language models (LLMs) have achieved remarkable results across various natural language processing applications \citep{NEURIPS2020_1457c0d6,wei2022chain,ouyang2022training,openai2024gpt4technicalreport}. This success largely depends on scaling the number of model parameters, the amount of training data, and computational resources \citep{kaplan2020scalinglawsneurallanguage,NEURIPS2022_c1e2faff}, which leads to substantial training and inference costs of LLMs. Building and deploying high-performance models also require enormous resources, posing a significant barrier for many researchers and practitioners.

%
% MoE
%
The \emph{Mixture of Experts} (MoE) architecture has emerged as a promising approach to address the escalating resource demands of LLMs. MoE introduces multiple experts into some parts of the network, but only a subset is activated at any given time, allowing the model to achieve superior performance with reduced training and inference costs \citep{shazeer2017,lepikhin2021gshard,Fedus2021SwitchTS}. In fact, cutting-edge industry models like Gemini 1.5 \citep{geminiteam2024gemini15unlockingmultimodal} and GPT-4 (based on unofficial reports) \citep{openai2024gpt4technicalreport} have adopted MoE, suggesting its effectiveness.

%
% MoE Challenge
%
We refer to transformer-based LLMs without MoE as \emph{dense models} and those incorporating MoE as \emph{MoE models}.
Upcycling~\citep{komatsuzaki2023sparse} is an approach that initializes and trains an MoE model using a pre-trained dense model, which aims to transfer learned knowledge for better initial performance.
However, \NUname{} copies the feedforward network (FFN) layers during initialization, which makes it difficult to achieve expert specialization.
This disadvantage prevents effective utilization of the MoE models' full capacity, resulting in slower convergence over long training periods.
Thus, there exists a trade-off between the short-term cost savings from knowledge transfer and the long-term convergence efficiency through expert specialization.

\begin{figure}[t]
    \centering
    \includegraphics[width=\textwidth]{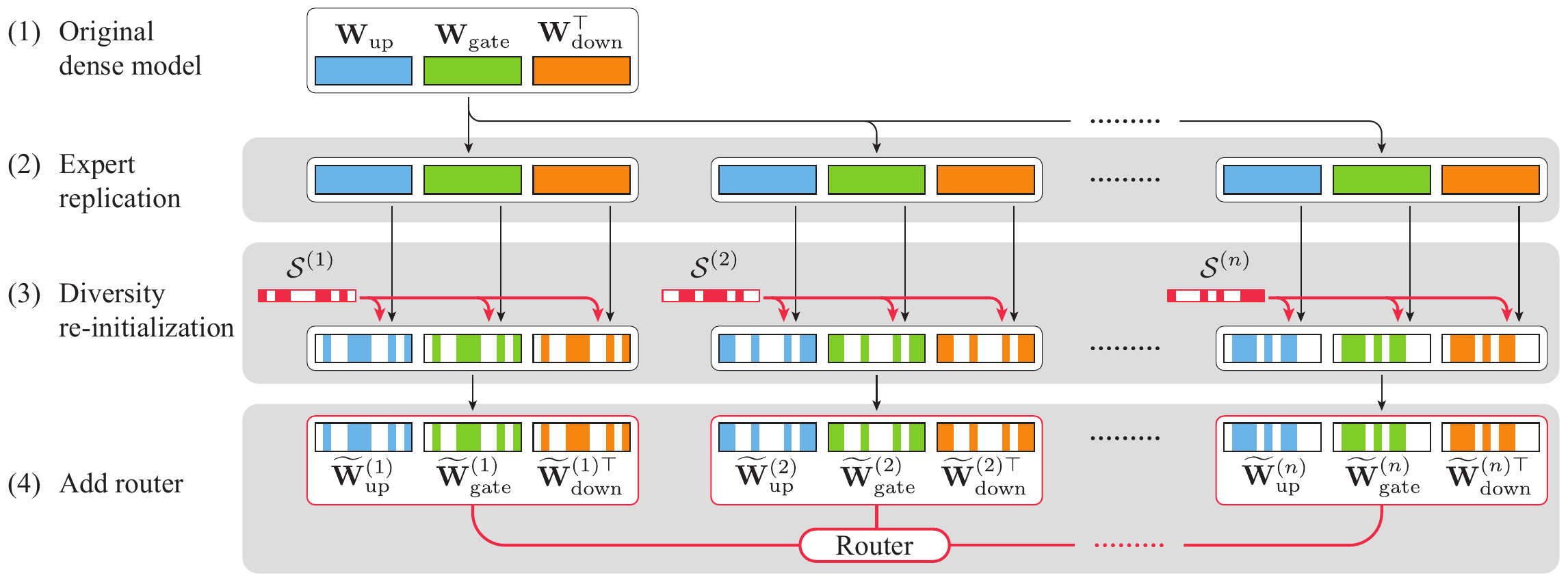}
\vskip -8pt 
\caption{\textbf{Overview of the \methodname{} method.} The key difference from the na\"{i}ve Upcycling is Diversity re-initialization, introduced in Section \ref{sec:method}.}
    \label{fig:drop_upcycling}
\end{figure}

In this paper, we propose \emph{\methodname{}} -- a method that effectively addresses this trade-off, as briefly illustrated in Figure \ref{fig:drop_upcycling}. \methodname{} works by selectively re-initializing the parameters of the expert FFNs when expanding a dense model into an MoE model. The method is carefully designed to promote expert specialization while preserving the knowledge of pre-trained dense models. Specifically, common indices are randomly sampled along the intermediate dimension of the FFNs, and the weights are dropped either column-wise or row-wise, depending on the weight matrix types. The dropped parameters are then re-initialized using the statistics of those weights.

%
% Experimental results
%
Extensive large-scale experiments demonstrate that \methodname{} nearly resolves the trade-off between the two aforementioned challenges
and significantly outperforms previous MoE model construction methods such as training from scratch and \NUname{}.
By leveraging pre-trained dense models, \methodname{} can start training from a better initial state than training from scratch, reducing training costs.
On the other hand, \methodname{} avoids the convergence slowdowns observed with \NUname{}.
Specifically, in our extensive long-term training experiments, \methodname{} maintained a learning curve slope similar to that of training from scratch, consistently staying ahead.
This success is attributed to effective expert specialization.
As a result, we constructed an MoE model with 5.9B active parameters that performs on par with a 13B dense model from the same model family, while requiring only approximately 1/4 of the training FLOPs.

%
% Fully open
%
This research is fully open, transparent, and accessible to all.
With over 200,000 GPU hours of experimental results, conducted on NVIDIA H100 GPUs, all training data, source code, configuration files, model checkpoints, and training logs used in this study are publicly available. By providing this comprehensive resource, we aim to promote further advancements in this line of research.

Our technical contributions are summarized as follows:
\begin{itemize}
\item We propose \methodname{}, a novel method for constructing MoE models that effectively balance knowledge transfer and expert specialization by selectively re-initializing parameters of expert FFNs when expanding a dense model into an MoE model.

\item Extensive large-scale experiments demonstrate that \methodname{} consistently outperforms previous MoE construction methods in long-term training scenarios.

\item All aspects of this research are publicly available. %, including training data, source code, configuration files, model checkpoints, and training logs. 
This includes the MoE model with 5.9B active parameters that performs comparably to a 13B dense model in the same model family while requiring only about 1/4 of the training FLOPs.

\end{itemize}

\section{Related Work}

\subsection{Mixture of Experts}
\label{sec:related_works:moe}

The concept of Mixture of Experts (MoE) was introduced about three decades ago~\citep{classic_moe_1,classic_moe_2}. Since then, the idea of using sparsely-gated MoE as a building block within neural network layers~\citep{moe_layer_iclr14,shazeer2017} has evolved and has been incorporated into transformer-based language models~\citep{lepikhin2021gshard, Fedus2021SwitchTS}. For a detailed overview of MoE, please refer to recent survey papers~\citep{moe_survey}.
Sparsely-gated MoE is currently the most common approach for building large-scale sparsely-activated models.
In this paper, we focus on sparsely-gated MoE (also referred to as sparse MoE or sparsely-activated MoE), and unless otherwise specified, the term MoE refers to it.

There are various designs of MoE layers and ways to integrate them into transformer-based LLMs. For example, in addition to the standard token-centric routing, expert-centric routing has also been proposed~\citep{expert_routing}. To incorporate common knowledge, it has been suggested to introduce shared experts that are always activated~\citep{dai-etal-2024-deepseekmoe}. To simplify the discussion, %unless otherwise specified, 
we assume the most standard top-$k$ token choice routing as the MoE layer and a decoder-only transformer-based LLM that uses MoE layers only in the FFNs as the MoE model. 
%This is because 
These are common design choices for recent MoE-based LLMs, such as Mixtral~\citep{jiang2024mixtralexperts}, Skywork-MoE~\citep{wei2024skyworkmoedeepdivetraining}, Phi-3.5-MoE~\citep{abdin2024phi3technicalreporthighly}, and Grok-1\footnote{\url{https://x.ai/blog/grok-os}}. 
% \citep{xai2024grok}
% \footnote{\url{https://x.ai/blog/grok-os}}. 
%
Specifically, these models use 8 experts (Mixtral and Grok-1) or 16 experts (Skywork and Phi-3.5-MoE), with the top-2 experts being activated per input token. Our experiments also use top-2 routing with 8 experts per layer, as this setup aligns with those practical configurations.
These facts indicate that \methodname{} can be applied to most variations of MoE models.
See Section~\ref{sec:methods:preliminaries} for technical details of MoE.

\subsection{MoE Model Initialization}
As with conventional neural networks, MoE models can be initialized randomly and trained from scratch. However, to reduce training costs, leveraging existing pre-trained dense models has become a standard approach. Below, we introduce a few methods for achieving this.

Upcycling~\citep{komatsuzaki2023sparse} leverages the weights of a pre-trained dense model for initializing an MoE model by initializing the experts in the MoE layer as replicas of the FFN layers in the dense model.
The main advantage of Upcycling is that it boosts the model's initial performance.  However, as our experiments show, MoE models initialized with Upcycling tend to have a much slower convergence, leading to suboptimal performance when trained for longer durations.

Branch-Train-MiX (BTX) \citep{sukhbaatar2024branchtrainmix} is a technique where a pre-trained dense model is replicated and fine-tuned on different datasets to produce multiple distinct expert dense models. These experts are then integrated into an MoE model, followed by additional training to optimize the routers. While this method appears to ensure expert specialization by design, \cite{jiang2024mixtralexperts} has highlighted that the diversity achieved in this way differs from that required for MoE layer experts, leading to suboptimal performance as a result. Our experiments also show that BTX suffers from suboptimal convergence similar to those observed in Upcycling.

Concurrent with our work, the Qwen2 technical report ~\citep{yang2024qwen2technicalreport} briefly suggests the use of a methodology possibly related to \methodname{} in training Qwen2-MoE. Due to the report's brevity and ambiguity, it is unclear if their method exactly matches ours. 
Our paper offers a valuable technical contribution even if the methods are similar. 
The potential application of \methodname{} in an advanced, industry-developed model like Qwen2-MoE that underscores the importance of further open investigation into this approach. We acknowledge the Qwen2 authors for sharing insights through their technical report.

\section{Method}
\label{sec:method}
In this section, we explain the \methodname{} method. \methodname{} initializes an MoE model by utilizing a pre-trained dense model and consists of three steps:

\begin{enumerate}
\item \textbf{Expert Replication:} The weights of the dense model are copied to create the MoE model. All layers, except for the FFN layers, are copied directly from the dense model. The FFN layers are replaced with MoE layers, and the original FFN weights are copied to all experts within these MoE layers.

\item \textbf{Diversity Re-initialization:} In each MoE layer, a subset of the expert parameters is randomly selected and re-initialized using the original statistical information. This promotes diversity among the experts while partially retaining the knowledge of the original model, which facilitates expert specialization during subsequent training.

\item \textbf{Continued Training:} After initialization, the MoE model is trained using the standard next-token prediction loss. Optionally, a load-balancing loss, commonly applied in MoE training, can also be incorporated.
\end{enumerate}
In the following, we explain the expert initialization and diversity injection processes.

\subsection{SwiGLU and MoE Layers}
\label{sec:methods:preliminaries}

We provide a brief overview of the MoE architecture. First, we review the feedforward network (FFN) layer in transformers. The SwiGLU activation function~\citep{shazeer2020gluvariantsimprovetransformer}, now standard in state-of-the-art LLMs like LLaMA~\citep{touvron2023llamaopenefficientfoundation} and Mixtral~\citep{jiang2024mixtralexperts}, will be used for explanation here. However, it should be noted that \methodname{} can be applied to transformers with any activation function. The FFN layer with SwiGLU is defined as follows:

\begin{equation}
\text{SwiGLU}(\mathbf{x}) = (\text{Swish}(\mathbf{x}^\mathrm{T} \mathbf{W}_\text{gate}) \odot \mathbf{x}^\mathrm{T} \mathbf{W}_\text{up}) \mathbf{W}_\text{down}.
\label{eq:ffn_swiglu}
\end{equation}
%ここで, $\mathbf{x} \in \mathbb{R}^{d_h}$は入力ベクトル, $\odot$はアダマール積を表し, 重み行列のサイズは以下の通りである：
Here, $ \mathbf{x} \in \mathbb{R}^{d_h}\ $ represents the input vector and \(\odot\) denotes the Hadamard product. Each FFN layer contains the following three weight matrices: $
%\begin{equation}
\mathbf{W}_\text{gate}, \mathbf{W}_\text{up} \in \mathbb{R}^{d_h \times d_f}$, and $\mathbf{W}_\text{down} \in \mathbb{R}^{d_f \times d_h}.
%\label{eq:weight_matrices}
%\end{equation}
$
The dimensions \(d_h\) and \(d_f\) are referred to as the hidden size and intermediate size, respectively.

When MoE is introduced into a transformer, each FFN layer is replaced with an MoE layer, while the rest of the architecture remains unchanged. Let us assume we use \(n\) experts and Top-$k$ gating. 
An MoE layer comprises a router and \(n\) expert FFNs. The router has a weight matrix \(\mathbf{W}_\text{router} \in \mathbb{R}^{d_h \times n}\). The $i$-th expert FFN is denoted as \(\text{SwiGLU}^{(i)}(\mathbf{x})\), which, like a standard FFN layer, consists of three weight matrices. These weights are denoted as \(\mathbf{W}^{(i)}_\text{gate}, \mathbf{W}^{(i)}_\text{up},\) and \(\mathbf{W}^{(i)}_\text{down}\). 
The output \(\mathbf{y}\) of the MoE layer is computed as follows:
%$n$エキスパートで, Top-Kゲーティングを使用する時, $\mathbf{W}_g \in \mathbb{R}^{d_h \times n}$をゲーティング重みとすると, MoE層の最終出力$\mathbf{y}$は以下のように計算される：
\begin{equation}
\mathbf{y} = \sum_{i=1}^{n} g(\mathbf{x})_i \cdot \text{SwiGLU}^{(i)}(\mathbf{x}),
\label{eq:moe_output}
\end{equation}
where \(g(\mathbf{x})_i\) is the $i$-th element of the output $g(\mathbf{x}) \in \mathbb{R}^n$ of the Top-$k$ routing function, defined as:
% ここで, $g_i(\mathbf{x})$はTop-Kゲーティング関数の出力で, 以下のように定義される：
\begin{equation}
g(\mathbf{x}) = \text{Softmax}(\text{Top-}k(\mathbf{x}^\mathrm{T} \mathbf{W}_\text{router})).
\label{eq:gating_function}
\end{equation}

Since \(k < n\) is typically the standard setting, only the top-$k$ selected experts out of \(n\) are computed. Therefore, the MoE layer is sparsely activated, meaning that only a subset of the parameters is involved in the computation. The number of parameters engaged in the computation for a given input is referred to as the \emph{active parameters} of the MoE model. This value is widely used as an approximation for the computational cost as it correlates well with the cost of both training and inference.
For non-MoE models, the total number of parameters corresponds to the active parameters as all parameters are involved in every computation.

\subsection{Expert Replication}
%まず、通常のFFN layerを用いたtransfomerのweightをコピーし、MoE layerを用いたtransformerを構築します。上で説明した通り、FFN層以外は全く同じアーキテクチャなので、それらについてはweightをそのままコピーします。各FFN層はMoE層に置き換える必要があります。新しいMoEレイヤーは以下の方法で作ります。

%routerのweightである $\mathbf{W}_\text{router}$ はランダムに初期化します。 $n$個のexpertについては、元のFFNのweightをコピーします。即ち、 \(\mathbf{W}^{(i)}_\text{gate} = \mathbf{W}_\text{gate}, \mathbf{W}^{(i)}_\text{up} = \mathbf{W}_\text{up},\) and \(\mathbf{W}^{(i)}_\text{down} = \mathbf{W}_\text{down}\) とします。

Following \citep{komatsuzaki2023sparse}, we first construct a Transformer with MoE layers by replicating the weights from a pre-trained Transformer with standard FFN layers. As explained earlier, the architecture remains identical except the FFN layers, so we simply copy the weights of all non-FFN components. Each FFN layer needs to be replaced with an MoE layer, and the new MoE layers are constructed as follows:
The router weights \(\mathbf{W}_\text{router}\) are initialized randomly. For the \(n\) experts, the weights from the original FFN are copied, such that \(\mathbf{W}^{(i)}_\text{gate} = \mathbf{W}_\text{gate}, \mathbf{W}^{(i)}_\text{up} = \mathbf{W}_\text{up},\) and \(\mathbf{W}^{(i)}_\text{down} = \mathbf{W}_\text{down}\).
% \footnote{There has been a recent approach that uses fine-grained experts by reducing the FFN width of MoE models~\citep{dai-etal-2024-deepseekmoe}. \methodname{} can be applied in this context as well. In this scenario, expert replication is performed by splitting either the columns ($\mathbf{W}_\text{gate}$ and $\mathbf{W}_\text{up}$) or the rows ($\mathbf{W}_\text{down}$) of the original FFN, and subsequent steps can be carried out in the same manner. \diff{See Appendix~\ref{appendix:extensions} for a detailed discussion.}}

% \diff{There has been a recent approach that uses fine-grained experts by reducing the FFN width of MoE models~\citep{dai-etal-2024-deepseekmoe}. See Appendix~\ref{appendix:extensions} for a detailed discussion.}

\diff{\methodname{} can also be applied to fine-grained experts and shared experts~\citep{dai-etal-2024-deepseekmoe}. See Appendix~\ref{appendix:extensions} for details.}

%密モデルのFFN層を$n$個のMoEモデルのエキスパート層に初期化する. 各エキスパート$j \in \{1, \ldots, n\}$の重み$\mathbf{W}^{(j)}$は以下のように初期化される：
%\begin{equation}
%\begin{aligned}
%\mathbf{W}^{(j)}_{up} &= \mathbf{W}_{up:,\mathcal{S}_j}, \quad \mathbf{W}^{(j)}_{gate} = %\mathbf{W}_{gate:,\mathcal{S}_j} \in \mathbb{R}^{d_h \times |\mathcal{S}_j|} \\
%\mathbf{W}^{(j)}_{down} &= \mathbf{W}_{down: ,\mathcal{S}_j,} \in \mathbb{R}^{|\mathcal{S}_j| \times d_h}
%\end{aligned}
%\label{eq:expert_init}
%\end{equation}
%ここで, $\mathcal{S}_j$は$\{1, \ldots, d_f\}$からランダムに選択された要素からなる集合, $|\mathcal{S}_j|$はその集合の要素数, $d_h$はモデル次元, $d_f$は元のFFN層の中間次元を表す. 

%本研究の実験では, $|\mathcal{S}_j| = d_f$ としている.

\subsubsection{Diversity Re-initialization}
\label{sec:reinit}

\begin{figure}[t]
\centering
\includegraphics[width=\textwidth]{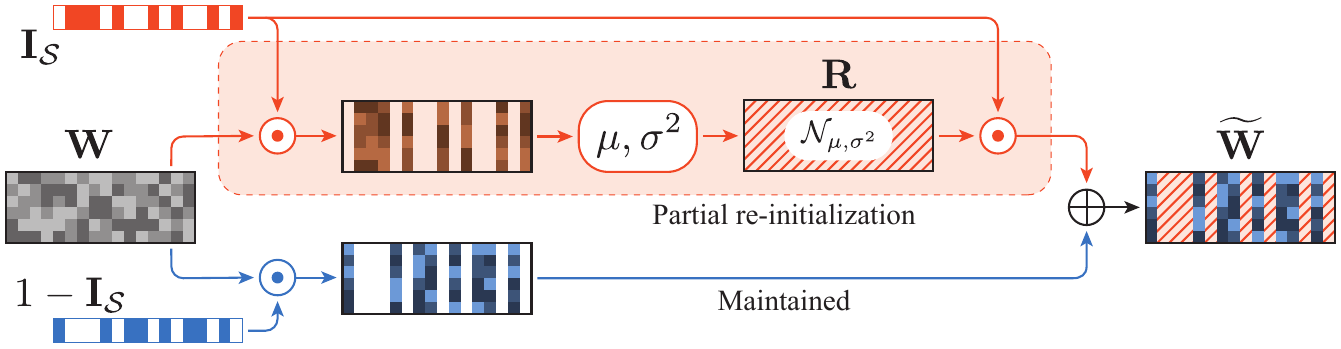}
\vskip -8pt
\caption{\textbf{Initialization of expert weights.} Columns (rows) are selected according to \diff{a set of randomly selected indices of the intermediate layer} $\mathcal{S}$, then all elements of them are re-initialized with the normal distribution. Other columns (rows) are maintained.}
\label{fig:partial-re-initialization}
\end{figure}

Diversity re-initialization is the key step in \methodname{}. 
%Re-initializing each expert randomly encourages the diversification of experts during subsequent training. 
This process is carefully designed to balance between knowledge retention and expert diversification. In particular, it is crucial to drop original weights along the intermediate dimension of the FFN layer based on shared indices across all three weight matrices. Specifically, the following operation is applied to every expert FFN in every MoE layer.

\paragraph{Step 1: Column-wise Sampling.}
We sample indices from the set of integers from 1 to intermediate size \(d_f\), namely,  $\mathcal{I}_{d_f}=\{ 1, 2, \cdots, d_f \}$, to create a set of partial indices \(\mathcal{S}\). A hyperparameter $r$ ($0 \leq r \leq 1$) controls the intensity of re-initialization, determining the proportion \(r\) used for sampling. That is, $\mathcal{S} \subseteq \mathcal{I}_{d_f}$ and $\left| \mathcal{S} \right| = \lfloor r d_f \rfloor$.

\paragraph{Step 2: Statistics Calculation.}  
We calculate the mean and standard deviation of the matrices of the weights corresponding to the selected indices $\mathcal{S}$. Specifically, we compute the mean and variance \((\mu_\text{up}, \sigma_\text{up})\), \((\mu_\text{gate}, \sigma_\text{gate})\), and \((\mu_\text{down}, \sigma_\text{down})\) 
from the values obtained only from the non-zero columns of $\mathbf{I}_{\mathcal{S}}$ in the products 
$\mathbf{I}_{\mathcal{S}}\odot W_{\text{gate}}$,
$\mathbf{I}_{\mathcal{S}} \odot W_{\text{up}}$, and 
$\mathbf{I}_{\mathcal{S}} \odot W_{\text{down}}^\top$, respectively, where $\mathbf{I}_{\mathcal{S}}$ is the indicator matrix whose values are 1 in the $i$-th column for $i\in\mathcal{S}$ and 0 otherwise.

%for sub-matrices \(W_{\text{up\ } :,\mathcal{S}}\), \(W_{\text{gate\ } :,\mathcal{S}}\), and \(W_{\text{down\ } \mathcal{S},:}\).

\paragraph{Step 3: Partial Re-Initialization.}
%最後に、これらを用いて3つの重み行列 $\mathbf{W}_\text{gate}$, $\mathbf{W}_\text{up}$, and $\mathbf{W}_\text{down}$ の部分的な最初期化をし、 $\widetilde{\mathbf{W}}_\text{gate}$, $\widetilde{\mathbf{W}}_\text{up}$, and $\widetilde{\mathbf{W}}_\text{down}$ を得る。
%選択されたindexについてはランダム初期化を、そうでないindexについては元のweightをコピーする。即ち、以下のようにする。
Finally, using the calculated statistics, we perform partial re-initialization of the three weight matrices \(\mathbf{W}_\text{gate}\), \(\mathbf{W}_\text{up}\), and \(\mathbf{W}_\text{down}\), obtaining \(\widetilde{\mathbf{W}}_\text{gate}\), \(\widetilde{\mathbf{W}}_\text{up}\), and \(\widetilde{\mathbf{W}}_\text{down}\). 
For the selected indices, the weights are dropped and re-initialized randomly, while for the unselected indices, the original weights are retained. 

%Let $\bar{\mathcal{S}}$ denotes the difference set of $\mathcal{S}$, that is, $\bar{\mathcal{S}} = \mathcal{I}_{d_f}\backslash\mathcal{S} $.
%
Let ${\mathbf{R}}_{\text{type}}$ be a matrix whose values are sampled from the $\mathcal{N}( \mu_{\text{type}}, ( \sigma_{\text{type}} )^2 )$ distribution, where type is one of the gate, up, or down, i.e., $\text{type} =\{\text{gate},\text{up},\text{down}\}$.
We then obtain $\widetilde{\mathbf{W}}_{\text{type}}$ by using the following equation:
\begin{equation}
\widetilde{\mathbf{W}}_{\text{type}} = \mathbf{I}_{\mathcal{S}} \odot \mathbf{R}_{\text{type}} +  (1 - \mathbf{I}_{\mathcal{S}}) \odot \mathbf{W}_{\text{type}}
,
\label{eq:DU_rand_init}
\end{equation}
where we consider that the matrices, $\widetilde{\mathbf{W}}_{\text{type}}$, ${\mathbf{R}}_{\text{type}}$, ${\mathbf{W}}_{\text{type}}$ are all transposed if $\text{type} = \text{down}$.

Figure~\ref{fig:partial-re-initialization} illustrates how we generate a single expert weight matrix from the original dense weights.

\subsubsection{Theoretical Characteristics}
Applying the re-initialization strategy explained above, the initial MoE model obtained by \methodname{} has the following characteristics:
\begin{enumerate}
\item \textbf{Parameter sharing among experts}:
since each expert retains the original representations with a ratio $(1-r)$, \diff{with Top-k routing where $k$ experts are selected, approximately $(1-r)^k$ of representations are preserved. }
\item \textbf{Characteristics of initial feedforward layers}:
\diff{Consider the output of an MoE layer with parameter re-initialization ratio $r$:}
% \begin{equation}
% \mathbf{y} \approx \text{FFN}_{\text{common}}(\mathbf{x}) + \sum_{i=1}^N g_i(\mathbf{x}) \cdot [\text{FFN}_{\text{retained}_i}(\mathbf{x}) - \text{FFN}_{\text{common}}(\mathbf{x}) + \text{FFN}_{\text{diverse}_i}(\mathbf{x})]
% \end{equation}
\begin{equation}
\diff{\mathbf{y} = \text{FFN}_{\text{common}}(\mathbf{x}) + \sum_{i=1}^N 
g(\mathbf{x})_i \cdot [\text{FFN}_{\text{retained}_i}(\mathbf{x}) - \text{FFN}_{\text{common}}(\mathbf{x}) + \text{FFN}_{\text{diverse}_i}(\mathbf{x})]}
\end{equation}
% where $\text{FFN}_{\text{common}}$ represents the output from parameters that are common to all selected $k$ experts (approximately ratio $(1-r)^k$ due to each expert independently preserving a ratio $(1-r)$ of original parameters), 
% $\text{FFN}_{\text{retained}_i}$ is expert $i$'s output using uniquely retained original parameters (ratio $(1-r)$), and $\text{FFN}_{\text{diverse}_i}$ is the output using reinitialized parameters (ratio $r$). The approximation error comes from parameter overlap, with magnitude $O(\frac{1}{\sqrt{d_f}})$. A detailed derivation is provided in Appendix~\ref{subsec:theoretical}.
% \end{enumerate}
\diff{where $\text{FFN}_{\text{common}}$ represents the output from parameters that are common to all selected $k$ experts (the proportion of such parameters is approximately $(1-r)^k$ due to each expert independently preserving a ratio $(1-r)$ of original parameters), 
$\text{FFN}_{\text{retained}_i}$ is expert $i$'s output using uniquely retained original parameters (ratio $(1-r)$), and $\text{FFN}_{\text{diverse}_i}$ is the output using reinitialized parameters (ratio $r$). The estimation error in the number of common parameters has magnitude $O\big(\frac{1}{\sqrt{\smash[b]{d_f}}}\big)$. A detailed derivation is provided in Appendix~\ref{subsec:theoretical}.}
\end{enumerate}

\section{Experimental Setup}

We conducted experiments to demonstrate the effectiveness of \methodname{} described in Section~\ref{sec:method}.
To clarify our model configurations, we introduce a notation where, for example, ``8×152M'' denotes an MoE model with eight experts and whose base dense model size is 152M.
%The following subsections explain the settings of our experiments.

% \subsection{Configuration of Models to compare}
\label{sec:model-architecture}

We selected the Llama~\citep{touvron2023llamaopenefficientfoundation} and Mixtral~\citep{jiang2024mixtralexperts} architectures for dense and MoE models, respectively, for our experiments. 
%%
%Both architectures are based on the Transformer~\citep{NIPS2017_3f5ee243} with several improvements, including RMSNorm~\citep{zhang-sennrich-neurips19}, SwiGLU~\citep{shazeer2020gluvariantsimprovetransformer}, and rotary position embeddings (RoPE)~\citep{su2024roformer}. 
%The notable difference in Mixtral (MoE) from Llama (dense) is that all feedforward network (FFN) layers are replaced by sparsely gated MoE layers.
%%
%
% We employed 8 experts and the dropless token choice Top-2 routing~\citep{megablocks} for the MoE. 
We employed 8 experts and the dropless~\citep{megablocks} token choice top-2 routing ~\citep{shazeer2017} for the MoE.
%Table~\ref{tab:model-details} shows the remaining hyper-parameters of the Dense and MoE models used in our experiments.
Detailed descriptions of the model configurations are provided in Appendix~\ref{appendix:model_configs_details}

We evaluated \diff{four} different methods to build MoE models, namely, training from scratch, \NUname{}~\citep{komatsuzaki2023sparse}, 
\diff{\RNUname{}~\citep{komatsuzaki2023sparse}} and Branch-Train-MiX~\citep{sukhbaatar2024branchtrainmix} to compare the performance with \methodname{}.
Moreover, we also evaluated dense models to provide a reference of the typical performance of LLMs in our configuration and illustrate the performance gains of MoE models.
We initialized all parameters of dense models using a Gaussian distribution $\mathcal{N}(0, 0.02)$.
The dense models are also used as the seed models of MoE models, except when we train MoE models from scratch.
When training MoE models from scratch, we used the same initialization method as the dense models, that is, $\mathcal{N}(0, 0.02)$.
%
%\textbf{MoE NU} is similar to \textbf{MoE DU}, but copies all parameters in \textbf{Dense} and duplicates FFN layers 8 times to initialize the MoE layers.
%
%We also evaluated Branch-Train-Mix proposed in~\citet{sukhbaatar2024branchtrainmix}.
\diff{In \RNUname{}, Drawing from \citep{muennighoff2024olmoeopenmixtureofexpertslanguage}, we initialize by copying the dense model parameters and then add Gaussian noise $\mathcal{N}(0, 0.02)$ to 50\% of the weights in each FFN layer.}
In Branch-Train-Mix, we first obtained three distinct expert dense models by further training a seed dense model with 100B extra tokens of either Japanese, English, or code. 
Then, we used the four dense models (the seed dense model and three expert dense models) to initialize the parameters of an MoE model.
Specifically, we averaged all parameters in the four dense models except the FFN layers and duplicated the FFN layers in each model twice to build eight MoE experts.
Note that this method involved extra training steps with 300B more tokens compared to the other MoE construction methods.
% Unless otherwise stated, dense models were trained on 1T tokens, and MoE models were trained on 500B tokens.

% \subsection{Training and Evaluation}

Unless otherwise stated, dense models were trained on 1T tokens, and MoE models were trained on 500B tokens.
Our training data was obtained from publicly available data.
We describe the detailed statistics of the training datasets in Appendix~\ref{appendix:dataset-details}.
%
% For dense models, we trained 1T tokens for the 1.5B model and 2.072T tokens for the 152M, 3.7B, and 13B models.
% Moreover, we trained 500B tokens for all MoE models.
% We used dense models trained on 1T tokens as the base for upcycling our MoE models. 
% For all MoE models except those trained from scratch, we Upcycled from 1T tokens trained dense checkpoints of all sizes.
%
We followed the typical training configurations used in Llama to train dense models and Mixtral for MoE models.
Details of the hyper-parameters we used are described in Appendix~\ref{appendix:training_configs_details}.
Moreover, the implementation and the computational environment used in our experiments are described in Appendix~\ref{appendix:training_environment}.

We conducted a comprehensive evaluation using a wide range of tasks in Japanese and English. 
We used 12 evaluation datasets that can be categorized into seven types.
The details of the evaluation datasets and metrics are described in Appendix \ref{appendix:evaluation_details}.

\section{Results and Discussion}
\label{sec:exp}

\begin{figure}[t]
\centering
\includegraphics[width=\textwidth]{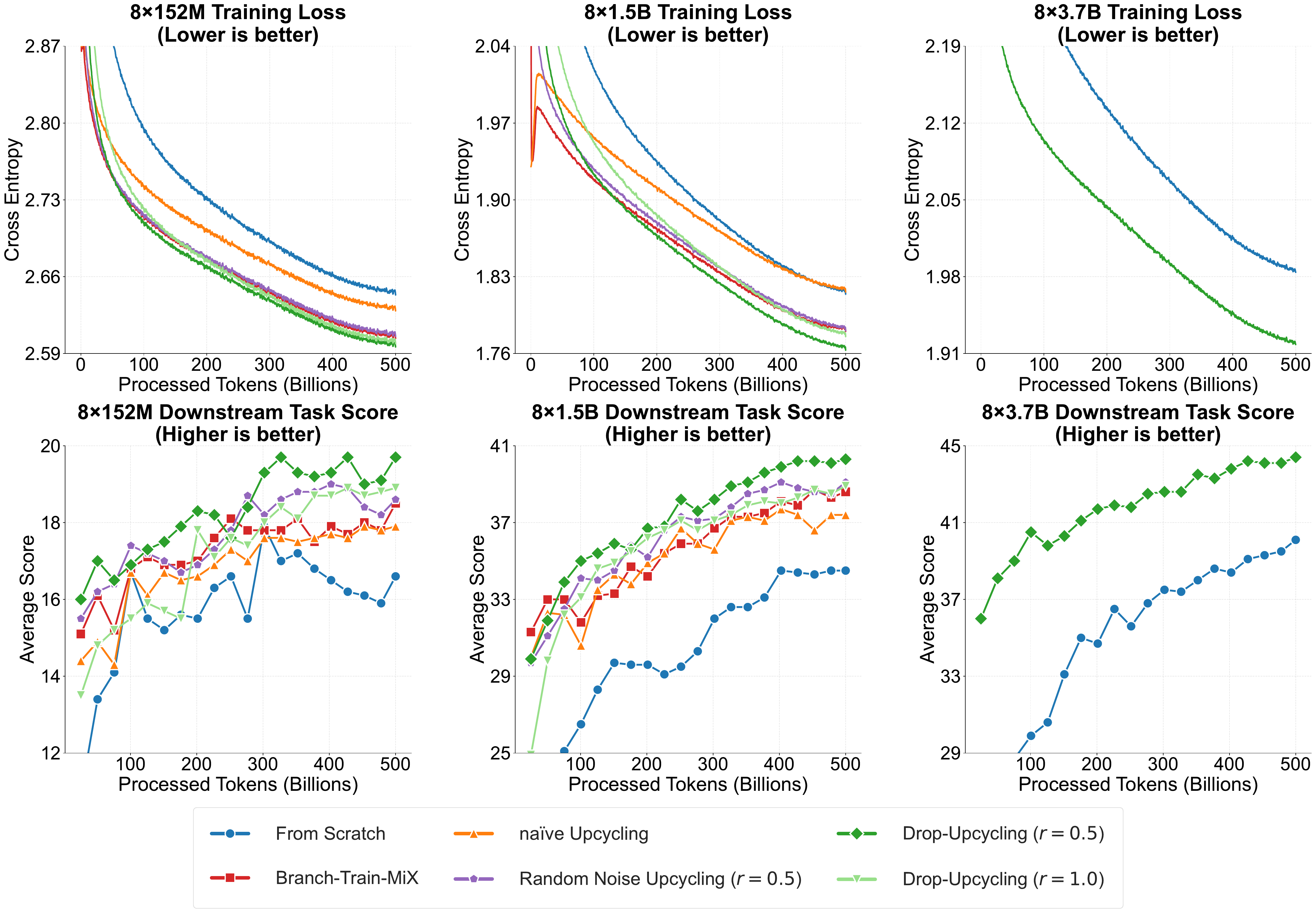} 
% \caption{\textbf{Evolution of Training Loss and Average Task Score for Different Model Sizes and Initialization Methods. }
% }
\vspace{-1em}
\caption{
\textbf{Comparison of learning curves for different MoE construction methods}.
The top and bottom rows illustrate the changes in training loss and downstream task scores during training, respectively. In both metrics, the proposed method, \methodname{} with $r=0.5$, achieves the best performance, gaining initial knowledge transfer while avoiding convergence slowdown. 
}

\label{fig:combined_train_loss_and_score}
\end{figure}

In this section, we address the following questions through experiments: 
 Is \methodname{} superior to existing MoE construction methods, and does \methodname{} resolve the issue of slower convergence? (Section~\ref{sec:result-method-comparison})  Does it perform well even in large-scale settings? (Section~\ref{sec:resuilts-scaling})  What is the impact of the re-initialization ratio 
$r$? (Section~\ref{sec:exp:ratio}) How are the experts specialized? (Section~\ref{sec:exp:diversity})

\subsection{Method Comparison}
\label{sec:result-method-comparison}
First, we compare \methodname{} with existing methods using small (8$\times$152M) to medium (8$\times$1.5B) scale settings. The left two columns of Figure~\ref{fig:combined_train_loss_and_score} illustrate the learning curves under these settings. The top and bottom rows illustrate the changes in training loss and downstream task scores during training, respectively. Note that in LLM pretraining, training loss serves as a reliable performance indicator since the risk of overfitting is low.
The performance on downstream tasks is represented by the average score across 12 tasks, which is commonly used as the overall evaluation metric. A detailed breakdown will be discussed later in conjunction with Table~\ref{tab:detailed-ja-en-comparison_method}.

Figure~\ref{fig:combined_train_loss_and_score} shows that \methodname{} at \( r = 0.5 \) (green) is significantly more efficient compared to other methods. The top row shows the training loss, while the bottom row displays the evaluation scores using downstream tasks. In both metrics and for both model sizes, \methodname{} becomes the clear winner after some training. Notably, the slope of the learning curve, which indicates convergence rate, is superior. Furthermore, it can be observed that the slope of the learning curve is consistent with the case of training from scratch, suggesting that \methodname{} resolves the crucial challenge of balancing knowledge transfer and expert specialization in Upcycling. For further analysis on expert specialization, see Section~\ref{sec:exp:diversity}.

Among existing methods, \NUname{} exhibited the slowest loss reduction rate and improvement in task scores. Branch-Train-Mix, which starts MoE training after each expert has been trained for 100B steps on different domains such as Japanese, English, and code, initially shows an advantage over \NUname{} due to this favorable initialization. However, its long-term learning pace is on par with \NUname{}, and it is ultimately overtaken by \methodname{}. As an ablation study, we evaluated setting \( r = 1.0 \) in \methodname{}, in addition to the standard \( r = 0.5 \). This configuration involves random initialization of all FFNs while reusing weights for embeddings and self-attention layers. This configuration might seem inefficient at first glance. Nevertheless, our large-scale experiments reveal that even such a seemingly naive baseline can outperform \NUname{} in certain scenarios. For additional analysis on the impact of the \( r \) value, refer to Section~\ref{sec:exp:ratio}.

% Table \ref{tab:detailed-ja-en-comparison_method} provides a comparison of the final downstream task performance for models trained with various methods under these 8$\times$152M and 8$\times$1.5B settings.
Table~\ref{tab:detailed-ja-en-comparison_method} provides a comparison of the final downstream task performance for models trained with various methods under these 8$\times$152M and 8$\times$1.5B settings. \diff{Model numbers refer to the leftmost column of this table.}
This table also includes the dense models used for upcycling. Specifically, Model 1 is the dense model used to initialize Models 3-\diff{7}, and Model \diff{8} is used to initialize Models \diff{10}-\diff{14}.
The proposed method, \methodname{} (DU) with $r=0.5$, consistently demonstrates superior performance across these model scales.
% In particular, both variants of DU (50\%) and DU (100\%) achieved results that outperform other approaches such as From Scratch (FS), \NUname{} (NU), and Branch-Train-MiX (BTX).

\begin{table}[t]
\caption{
\textbf{Comparison of evaluation results between models with different initialization.} 
Training from scratch (FS), Branch-Train-Mix (BTX), \NUname{} (NU), \diff{\RNUname{} (RNU)} and \methodname{} (DU) are compared. 
\diff{$^*$ BTX requires additional 300B tokens to obtain specialized dense models before MoE construction.}
%In addition to the individual scores for each downstream task, we also present the average score across 12 tasks, which is commonly used as the overall evaluation metric for the models.
%For both individual scores and the average, higher values indicate better performance.
Bold letters indicate the highest score within each model size.
%\textbf{Performance comparison of downstream tasks across different model sizes and training methods.} Scores represent task-specific metrics (higher is better). Bold indicates the best score for each model size.
}

\label{tab:detailed-ja-en-comparison_method}
\centering
\small
\renewcommand{\arraystretch}{1.03}
% \tabcolsep=0.15cm
% \begin{adjustbox}{width=\linewidth}
% \begin{tabular}{cll*{14}{r}}
% \toprule
% & \multicolumn{2}{c}{\textbf{Model}} & & \multicolumn{13}{c}{\textbf{Individual Scores}} \\
% \cmidrule{2-3} \cmidrule{5-16}
% \textbf{\#} &
% \makecell[c]{\textbf{Archi-} \\ \textbf{tecture}} & \makecell[c]{\textbf{MoE} \\ \textbf{Init}} & & \makecell[c]{\textbf{JEM}\\\textbf{HQA}} & \makecell[c]{\textbf{NIILC}} & \makecell[c]{\textbf{JSQ}} & \makecell[c]{\textbf{XL}-\\\textbf{Sum}} & \makecell[c]{\textbf{WMT}\\\textbf{E$\to$J}} & \makecell[c]{\textbf{WMT}\\\textbf{J$\to$E}} & \makecell[c]{\textbf{OB}\\\textbf{QA}} & \makecell[c]{\textbf{TQA}} & \makecell[c]{\textbf{HS}} & \makecell[c]{\textbf{SQ}\\\textbf{v2}} & \makecell[c]{\textbf{XW}-\\\textbf{EN}} & \makecell[c]{\textbf{BBH}} & \makecell[c]{\textbf{Avg}} \\

\tabcolsep=0.12cm
\begin{adjustbox}{width=\linewidth}
\begin{tabular}{cllrrrr*{13}{r}}
% \begin{tabular}{cllrrrrrr*{13}{r}}
\toprule
% & \multicolumn{2}{c}{\textbf{Model}} & \multicolumn{2}{c}{\textbf{Training}} & \multicolumn{13}{c}{\textbf{Individual Scores}} \\
% \cmidrule{2-3} \cmidrule{4-5} \cmidrule{6-17}
& \multicolumn{2}{c}{\textbf{Model}} & & \multicolumn{2}{c}{\diff{\textbf{Training}}} & & \multicolumn{13}{c}{\textbf{Individual Scores}} \\
\cmidrule{2-3} \cmidrule{5-6} \cmidrule{8-19}

\textbf{\#} &
\makecell[c]{\textbf{Archi-} \\ \textbf{tecture}} & \makecell[c]{\textbf{MoE} \\ \textbf{Init}} & & \makecell[c]{\diff{\textbf{Tokens}}} & \makecell[c]{\diff{\textbf{FLOPs}} \\ \diff{($\times 10^{21}$)}} & & \makecell[c]{\textbf{JEM}\\\textbf{HQA}} & \makecell[c]{\textbf{NIILC}} & \makecell[c]{\textbf{JSQ}} & \makecell[c]{\textbf{XL}-\\\textbf{Sum}} & \makecell[c]{\textbf{WMT}\\\textbf{E$\to$J}} & \makecell[c]{\textbf{WMT}\\\textbf{J$\to$E}} & \makecell[c]{\textbf{OB}\\\textbf{QA}} & \makecell[c]{\textbf{TQA}} & \makecell[c]{\textbf{HS}} & \makecell[c]{\textbf{SQ}\\\textbf{v2}} & \makecell[c]{\textbf{XW}-\\\textbf{EN}} & \makecell[c]{\textbf{BBH}} & \makecell[c]{\textbf{Avg}} \\

% \midrule
% & & 4shot & 4shot  & 4shot  & 1shot  & 4shot  & 4shot  & 4shot  & 4shot  & 4shot  & 4shot  & 4shot  & 3 shot (CoT)&\\
\midrule
% \multicolumn{4}{l}{\textbf{\textit{152Mベースでの手法比較実験:}}}  \\
\multicolumn{6}{l}{\textbf{\textit{Dense 152M $\to$ MoE 8$\times$152M:}}} \\
1 & Dense & -- & & \diff{1,000B} & \diff{1.59} & & 17.6 & 7.9 & 10.6 & 2.4 & 0.5 & 0.5 & 14.6 & 3.0 & 28.6 & 2.0 & 60.6 & 11.5 & 13.3 \\
2 & MoE  & FS & & \diff{500B} & \diff{0.91} & & 25.2	&13.6	&19.4	&1.8	&0.9	&0.4	&16.6	&2.6	&31.2	&\textbf{12.9}	&64.4	&10.7	&16.6 \\
3 & MoE &BTX & & \diff{800B$^*$} & \diff{1.39} & & 28.6	&17.1	&26.6	&\textbf{4.3}	&2.7	&1.1	&\textbf{18.4}	&5.1	&\textbf{32.5}	&5.3	&\textbf{65.0}	&15.9	&18.5 \\
4 &MoE & NU & & \diff{500B} & \diff{0.91} & & 28.2	&16.2	&24.4	&3.5	&3.0	&1.1	&18.2	&5.8	&31.9	&4.5	&63.5	&14.7	&17.9 \\
\diff{5} & \diff{MoE} & \diff{RNU ($r$=0.5)} & & \diff{500B} & \diff{0.91} & & \diff{28.6} & \diff{17.1} & \diff{29.4} & \diff{3.7}	& \diff{2.3} & \diff{1.6} & \diff{16.8} & \diff{5.3}	& \diff{32.0} & \diff{4.8} & \diff{64.5} & \diff{17.4} & \diff{18.6} \\
\rowcolor{verylightgray} \diff{6} & MoE & DU ($r$=0.5) & & \diff{500B} & \diff{0.91} & & \textbf{32.2}	&\textbf{18.0}	&30.6	&3.7	&\textbf{4.7}	&\textbf{2.3}	&16.8	&\textbf{6.1}	&\textbf{32.5}	&6.2	&64.2	&\textbf{19.1}	&\textbf{19.7} \\
\diff{7} &MoE & DU ($r$=1.0) & & \diff{500B} & \diff{0.91} & & 27.2	&16.8	&\textbf{32.5}	&4.1	&3.7	&1.6	&17.0	&5.9	&32.4	&4.9	&64.8	&15.4	&18.9 \\

 \midrule
%\multicolumn{4}{l}{\textbf{\textit{1.5B base model comparison:}}} \\
\multicolumn{6}{l}{\textbf{\textit{Dense 1.5B $\to$ MoE 8$\times$1.5B:}}} \\
\diff{8} & Dense & -- & & \diff{1,000B} & \diff{11.76} & & 49.6 & 42.5 & 48.1 & 11.3 & 16.8 & 8.5 & 22.2 & 23.8 & 42.9 & 16.2 & 82.5 & 25.1 & 32.5 \\
\diff{9} & MoE & FS & & \diff{500B} & \diff{9.05} & & 48.3 & 45.4 & 59.1 & 7.5 & 16.6 & 6.9 & 26.4 & 31.5 & 47.3 & 15.0 & 83.7 & 25.9 & 34.5 \\
\diff{10} & MoE & BTX & & \diff{800B$^*$} & \diff{12.58} & & 44.3	&51.8	&69.4	&11.9	&22.4	&\textbf{12.5}	&27.8	&39.2	&49.7	&18.7	&\textbf{86.4}	&28.9	&38.6\\
\diff{11} & MoE & NU & & \diff{500B} & \diff{9.05} & & 50.4 & 50.6 & 61.7 & 12.4 & 21.6 & 10.5 & 26.8 & 36.2 & 47.7 & 19.0 & 85.0 & 27.2 & 37.4 \\
\diff{12} & \diff{MoE} & \diff{RNU ($r$=0.5)} & & \diff{500B} & \diff{9.05} & & \diff{\textbf{53.6}} & \diff{50.5} & \diff{71.2} & \diff{12.3} & \diff{22.3}& \diff{11.7} & \diff{26.4} & \diff{40.0} & \diff{49.9} & \diff{19.1} & \diff{84.9} & \diff{27.5} & \diff{39.1} \\
 \rowcolor{verylightgray} \diff{13} & MoE & DU ($r$=0.5) & & \diff{500B} & \diff{9.05} & & 51.1 & \textbf{52.3} & \textbf{72.5} & \textbf{13.7} & \textbf{22.5} & \textbf{12.5} & \textbf{30.6} & \textbf{41.3} & \textbf{50.4} & \textbf{21.2} & 86.2 & \textbf{29.1} & \textbf{40.3} \\
\diff{14} & MoE & DU ($r$=1.0) & & \diff{500B} & \diff{9.05} & & 52.1 & 50.9 & 68.8 & 12.3 & 21.9 & 12.4 & 25.0 & 39.1 & 49.7 & 20.6 & 86.0 & 27.9 & 38.9 \\
%  \midrule
% \multicolumn{4}{l}{\textbf{\textit{3.7Bベースでのスケーリングの実験: }}}  \\
% Dense$^1$ & -- & 44.5 & 47.2 & 78.8 & 12.8 & 21.4 & 15.4 & 25.0 & 33.8 & 47.3 & 23.7 & 85.9 & 28.7 & 38.7 \\
% \multirow{2}{*}{8$\times$3.7B} & FS &53.5	&50.8	&69.6	&10.4	&20.6	&13.9	&29.0	&45.8	&51.1	&21.1	&87.1	&28.1	&40.1 \\
%  & DU (50\%) &
%  \textbf{47.5}&	\textbf{57.0}	&\textbf{82.2}&	\textbf{16.3}	&\textbf{25.0}&	\textbf{19.0}&	\textbf{31.2}&	\textbf{53.6}	&\textbf{54.4}&	\textbf{26.3}&	\textbf{88.5}&	\textbf{32.2}&	\textbf{44.4}\\
% \multicolumn{4}{l}{\textbf{\textit{3.7Bベースでのスケーリングの実験のベースライン: }}}  \\
% 13B Dense& FS &47.6	&58.3	&85.2	&14.1	&24.6	&18.3	&31.4	&48.6	&53.1	&29.3	&88.3	&35.2	&44.5 \\

% 3.7B Dense (2.1T)& FS &42.3	&53.2	&80.4	&14.3	&22.6	&15.9	&28.2	&42.2	&50.6	&25.8	&87.3	&30.9	&41.1 \\

\bottomrule
\end{tabular}
\end{adjustbox}
%\raggedright
%\small
%$^1$ Checkpoint prior to upcycling to MoE. \\
\end{table}

\begin{table}[t]
\caption{\textbf{Comparison between dense and MoE with large-scale configuration.}
\methodname{} (DU) works well even at 8$\times$3.7B scale. The MoE model with \methodname{} outperforms dense models trained with higher computational costs, demonstrating the effectiveness of \methodname{}.
%Model 1 is a dense model created to be upcycled, and Models 2 and 3 are MoE models constructed using different methods. In addition, for comparison, we include Models 4 and 5, which are dense models built with more computational resources using the same methodology. Despite using fewer computational resources, the MoE model created through \methodname{} outperforms these dense models, demonstrating the effectiveness of \methodname{}.
%\caption{\textbf{Performance comparison of downstream tasks across different model sizes and training methods.} Scores represent task-specific metrics (higher is better). Bold indicates the highest score across all model sizes and methods.
}

\tabcolsep 3pt%=0.1cm
\label{tab:detailed-ja-en-comparison_method_scaling}
\centering
\small
\renewcommand{\arraystretch}{1.3}
\begin{adjustbox}{width=\linewidth}
\begin{tabular}{cllrrrrrr*{13}{r}}
\toprule

%Model & Training Method & A para& Para &  flops & tokens& \makecell[c]{JEM\\HopQA} & \makecell[c]{NIILC} & \makecell[c]{JSQuAD} & \makecell[c]{XL-\\Sum} & \makecell[c]{WMT20\\En$\to$Ja} & \makecell[c]{WMT20\\Ja$\to$En} & \makecell[c]{Open\\BookQA} & \makecell[c]{Triv.\\QA} & \makecell[c]{Hella\\Swag} & \makecell[c]{SQuAD\\v2} & \makecell[c]{XWino-\\grad EN} & \makecell[c]{BBH} & AVG \\
& \multicolumn{3}{c}{\textbf{Model}} & & \multicolumn{2}{c}{\textbf{Training}} & & \multicolumn{13}{c}{\textbf{Individual Scores}} \\
\cmidrule{2-4} \cmidrule{6-7} \cmidrule{9-20} 
% \rule{0pt}{4ex}
%\makecell[c]{\textbf{Architecture}} & \makecell[c]{\textbf{MoE} \\ \textbf{Init}} & \makecell[c]{\textbf{Act Params} / \\\textbf{Total Params}} & & \makecell[c]{\textbf{Tokens}} & \makecell[c]{\textbf{FLOPs}} & & \makecell[c]{\textbf{JEM}\\\textbf{HopQA}} & \makecell[c]{\textbf{NIILC}} & \makecell[c]{\textbf{JSQuAD}} & \makecell[c]{\textbf{XL}-\\\textbf{Sum}} & \makecell[c]{\textbf{WMT20}\\\textbf{En$\to$Ja}} & \makecell[c]{\textbf{WMT20}\\\textbf{Ja$\to$En}} & \makecell[c]{\textbf{Open}\\\textbf{BookQA}} & \makecell[c]{\textbf{Triv.}\\\textbf{QA}} & \makecell[c]{\textbf{Hella}\\\textbf{Swag}} & \makecell[c]{\textbf{SQuAD}\\\textbf{v2}} & \makecell[c]{\textbf{XWino}-\\\textbf{grad EN}} & \makecell[c]{\textbf{BBH}} & \textbf{Avg} \\
\makecell[c]{\textbf{\#}} &
\makecell[c]{\textbf{Architecture}} & \makecell[c]{\textbf{MoE} \\ \textbf{Init}} & \makecell[c]{\textbf{Act Params} / \\\textbf{Total Params}} & & \makecell[c]{\textbf{Tokens}} & \makecell[c]{\textbf{FLOPs} \\ ($\times 10^{22}$)} & & \makecell[c]{\textbf{JEM}\\\textbf{HQA}} & \makecell[c]{\textbf{NIILC}} & \makecell[c]{\textbf{JSQ}} & \makecell[c]{\textbf{XL}-\\\textbf{Sum}} & \makecell[c]{\textbf{WMT}\\\textbf{E$\to$J}} & \makecell[c]{\textbf{WMT}\\\textbf{J$\to$E}} & \makecell[c]{\textbf{OB}\\\textbf{QA}} & \makecell[c]{\textbf{TQA}} & \makecell[c]{\textbf{HS}} & \makecell[c]{\textbf{SQ}\\\textbf{v2}} & \makecell[c]{\textbf{XW}-\\\textbf{EN}} & \makecell[c]{\textbf{BBH}} & \textbf{Avg} \\

\midrule
%\multicolumn{6}{l}{\textbf{\textit{3.7Bベースでのスケーリングの実験: }}}  \\
1 & Dense 3.7B & - &  3.7B / 3.7B & & 1,000B& 2.70 & & 44.5 & 47.2 & 78.8 & 12.8 & 21.4 & 15.4 & 25.0 & 33.8 & 47.3 & 23.7 & 85.9 & 28.7 & 38.7 \\
2 & MoE 8$\times$3.7B & FS & 5.9B / 18B & & 500B & 1.98 & & \textbf{53.5}	&50.8	&69.6	&10.4	&20.6	&13.9	&29.0	&45.8	&51.1	&21.1	&87.1	&28.1	&40.1 \\
\rowcolor{verylightgray} 3 & MoE 8$\times$3.7B  & DU ($r$=0.5) &5.9B / 18B & & 500B & 1.98 & &
 47.5&	\textbf{57.0}	&82.2&	\textbf{16.3}	&\textbf{25.0}&	\textbf{19.0}&	31.2&	\textbf{53.6}	&\textbf{54.4}&	26.3&	\textbf{88.5}&	32.2&	44.4\\
%\multicolumn{6}{l}{\textbf{\textit{3.7Bベースでのスケーリングの実験のベースライン: }}}  \\
4 & Dense 13B& - &13B / 13B & & 805B & 7.43 & &47.6	&58.3	&\textbf{85.2}	&14.1	&24.6	&18.3	&\textbf{31.4}	&48.6	&53.1	&\textbf{29.3}	&88.3	&\textbf{35.2}	&\textbf{44.5} \\

5 & Dense 3.7B & -&3.7B / 3.7B & & 2,072B &5.58 & &42.3	&53.2	&80.4	&14.3	&22.6	&15.9	&28.2	&42.2	&50.6	&25.8	&87.3	&30.9	&41.1 \\

\bottomrule
\end{tabular}
\end{adjustbox}
%\raggedright
%\small
%$^1$ Checkpoint prior to upcycling to MoE. \\
\end{table}

% \subsection{Scaling to 8$\times$3.7 B}
\subsection{Scaling to \texorpdfstring{8$\times$3.7}{8x3.7}B}

\label{sec:resuilts-scaling}
To further evaluate the effectiveness of \methodname{} in larger-scale settings and to build a practical MoE model, we conducted experiments with an 8$\times$3.7B configuration. Due to computational resource constraints, experiments under the 8$\times$3.7B setting were limited to training from scratch and \methodname{} with \( r = 0.5 \).

The rightmost column of Figure~\ref{fig:combined_train_loss_and_score} illustrates the learning curves under this configuration. Similar to the 8$\times$152M and 8$\times$1.5B settings, \methodname{} significantly outperforms training from scratch. There is an initial gain in performance due to the improved initialization, and expert diversification allows the training to progress as efficiently as in the case of training from scratch, ensuring that \methodname{} never gets overtaken.

Table~\ref{tab:detailed-ja-en-comparison_method_scaling} compares the models' final downstream task performance. \diff{Model numbers refer to the leftmost column of this table.}
Model 1 is a dense model used as a base model for the Upcycling.
Models 2 and 3 are MoEs built using \NUname{} and \methodname{}, respectively, demonstrating the superiority of \methodname{}.
In addition, two different baseline dense models, Models 4 and 5, are included in the table. Model 4 is a 13B dense model. Our 8$\times$3.7B MoE architecture has fewer active parameters than this 13B model, leading to lower training and inference costs. Nevertheless, the 8$\times$3.7B MoE model using \methodname{} achieves better performance upon completion of training. Model 5 is a 3.7B dense model trained with 2.1T tokens. The fact that our 8$\times$3.7B MoE model with \methodname{} surpasses this dense model indicates that rather than continuously investing resources into training dense models, it might be a superior option to convert them to MoE models through \methodname{} and continue training at a certain point in the process.

\subsection{Analysis 1: Re-initializaiton Ratio}
\label{sec:exp:ratio}

\begin{figure}[t]
\centering
\includegraphics[width=\textwidth]
% {images/initialization_ratio_comparison.pdf}
{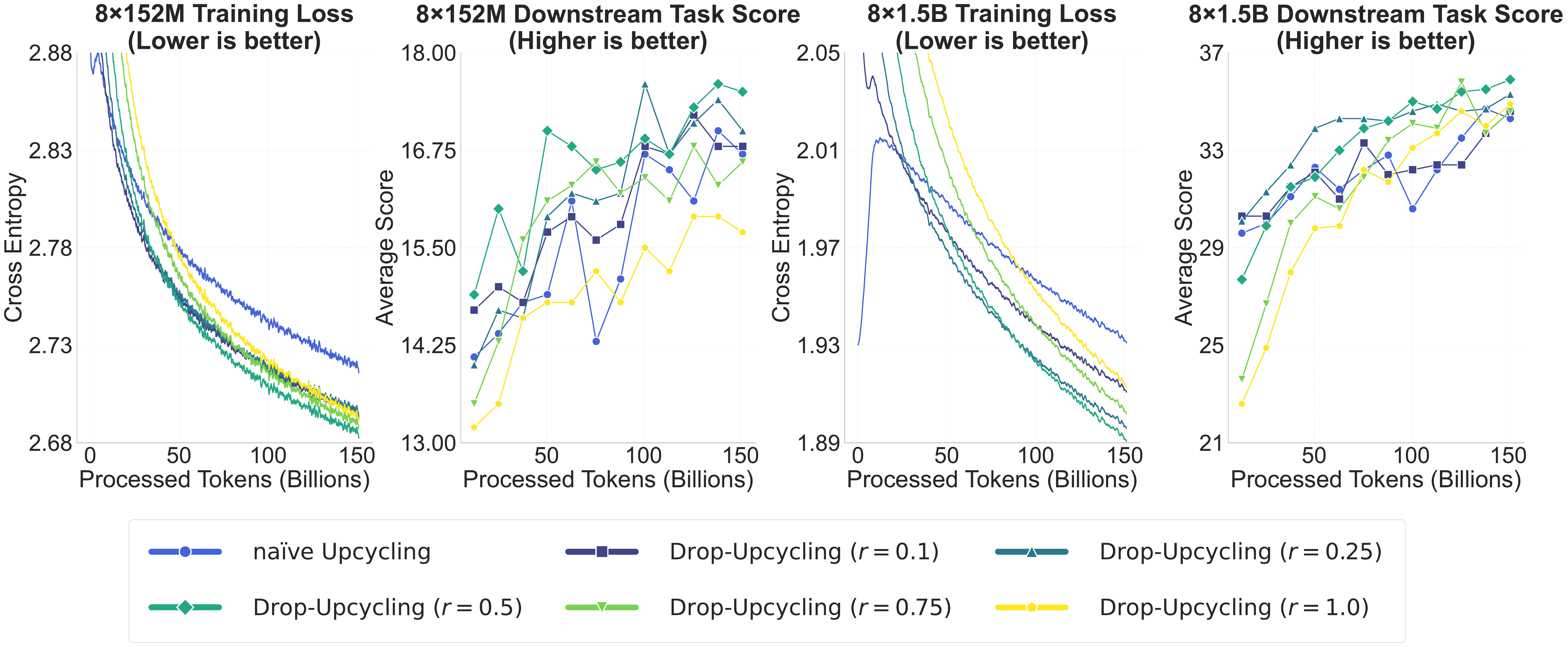}
\vskip -5pt 
\caption{\textbf{Impact of re-initialization ratio $r$.} The training loss and downstream task score over the total number of tokens processed during training on 8×152M (left two figures) and 8×1.5B (right two figures) settings are illustrated.
Even with different $r$ values, \methodname{} robustly outperforms \NUname{}, and 0.5 appears to be the most effective ratio.}
%Training Loss and averaged Downstream Task Score over the total numbers of tokens processed during training}

\label{fig:initialization_ratio_comparison}
\end{figure}

We conducted a study to investigate the impact of the re-initialization ratio $r$ in \methodname{}.
Figure \ref{fig:initialization_ratio_comparison} illustrates the effects of different re-initialization rates 0.0 (\NUname{}), 0.1, 0.25, 0.5, 0.75, and 1.0 on models of sizes 8×152M and 8×1.5B. Each model was trained up to 150B tokens, during which we monitored the training loss and the progression of the average downstream task scores.

The experimental results revealed similar trends across both model sizes. In terms of long-term performance, a re-initialization ratio of 0.5 yielded the best results for both models, maintaining superiority in both training loss and average task scores.
An interesting pattern emerged regarding the influence of the re-initialization ratio. With lower re-initialization rates, particularly at 0.0 (\NUname{}), the models struggled to significantly improve beyond the performance of the original pre-trained models. While re-initialization rates of 0.1 and 0.25 showed promising performance in the early stages of training, they were eventually surpassed by the 0.5 re-initialization rate as training progressed.
These observations suggest that increasing the re-initialization ratio helps the models escape local optima, enabling more effective learning. However, excessively high re-initialization rates of 0.75 or 1.0 appeared to hinder the effective knowledge transfer from the pre-trained dense models.
This phenomenon highlights an important trade-off concerning the MoE initialization: a balance must be struck between knowledge transfer and effective expert specialization.
\methodname{} with \( r = 0.5 \) is a robust and practical method that ideally balances these two aspects.

% Regarding the relationship between task scores and training loss, although some variability was observed in the task scores, the overall trend confirmed that as the training loss decreased (indicating training progression), the task scores improved.

\subsection{Analysis 2: Expert Specialization}
\label{sec:exp:diversity}

We analyze expert routing patterns to examine how \methodname{} facilitates expert specialization. We apply the methodologies of \citet{jiang2024mixtralexperts} and \citet{muennighoff2024olmoeopenmixtureofexpertslanguage} to 8×1.5B MoE models trained with different methods. This analysis investigates how data from different domains is routed to various experts. As input data from different domains, we use the validation sets from Japanese and English Wikipedia; the validation set of the Japanese MC4 dataset (as split by the authors; see \citealt{llmjp2024llmjpcrossorganizationalprojectresearch}), originally introduced by \citet{2019t5}; The Stack \citep{kocetkov2023the}; and the English C4 dataset \citep{muennighoff2024olmoeopenmixtureofexpertslanguage}.

\begin{figure}[t] \centering \includegraphics[width=\textwidth]{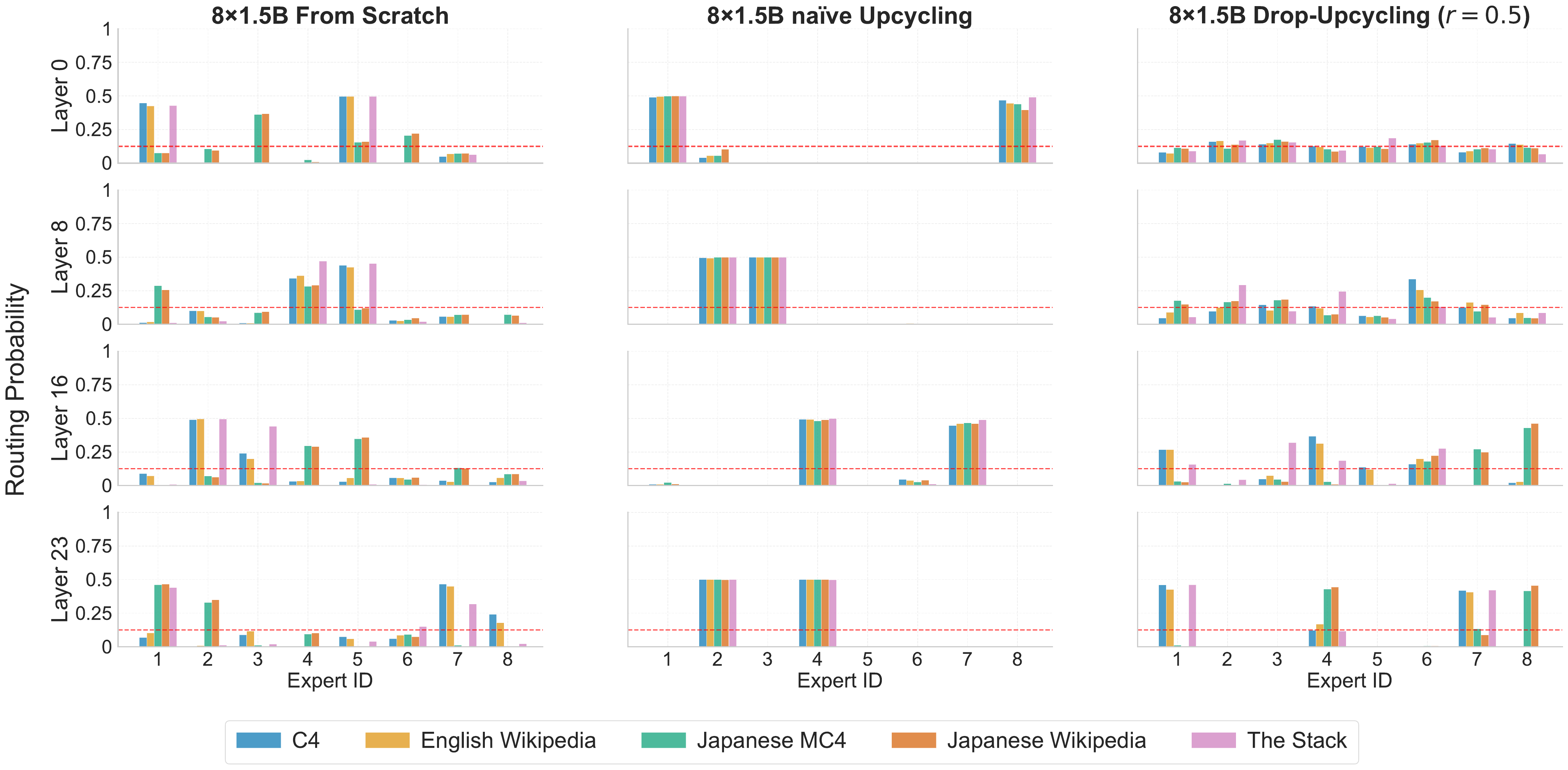} \vskip -3pt 
\caption{
\textbf{Comparison of expert routing patterns across different MoE construction methods.} 
\methodname{} exhibits more balanced expert utilization than \NUname{}.
Results shown for layers 0 (first), 8, 16, and 23 (last); see Appendix~\ref{subsec:detailed_routing_analysis} for results on all layers.
%\textbf{Expert routing patterns} at layers 0 (first), 8, 16, and 23 (last) on different datasets.
} \label{fig:routing} \end{figure}

In Figure~\ref{fig:routing}, we observe that \NUname{} \diff{with global load balancing} results in a highly imbalanced routing pattern, \diff{where} the majority of experts were underutilized or not utilized at all, with only two experts being always selected across all layers. \diff{While layer-wise load balancing mitigate such expert collapse, we found no significant differences in the training loss trajectories or model performance between these two strategies (see Appendix~\ref{subsec:detailed_load_balance_comparison}).} In contrast, both the model trained from scratch and the one enhanced with \methodname{} (with $r=0.5$) exhibit \diff{domain-specialized routing patterns regardless of the load balancing strategy.} The routing patterns reveal that certain experts specialize in processing specific types of data, such as Japanese text, English text, or code snippets\diff{, as} evident from the distinct expert selection probabilities corresponding to each dataset.

These findings suggest that \methodname{} promotes effective expert specialization \diff{independently of the load balancing strategy, which likely contributes to the improved performance observed in our experiments. For detailed routing patterns across all 24 layers and further analysis of load balancing strategies, see Appendix~\ref{subsec:detailed_routing_analysis} and \ref{subsec:detailed_load_balance_comparison}.}

\section{Conclusion}

In this paper, we introduced \methodname{}, a novel method for efficiently constructing Mixture of Experts (MoE) models from pre-trained dense models. Selectively re-initializing parameters of expert feedforward networks, \methodname{} 
 effectively balances knowledge transfer and expert specialization, addressing the key challenges in MoE model development.

Our extensive large-scale experiments demonstrated that \methodname{},    
significantly outperforms previous MoE construction methods. As a result, we achieved an MoE model with 5.9B active parameters that matches the performance of a 13B dense model from the same model family while requiring only about 1/4 of the training FLOPs.

By making all aspects of our research publicly available—including data, code, configurations, checkpoints, and logs—we aim to promote transparency and facilitate further advancements in efficient LLM training. We believe that \methodname offers a practical solution to reduce resource barriers in deploying high-performance LLMs, contributing to broader accessibility and innovation in AI research.

\ificlrfinal
\section*{Acknowledgements}

The authors would like to thank Masanori Suganuma and Kou Misaki for providing valuable discussions and feedback during the preparation of this manuscript.
This work was supported by the "R\&D Hub Aimed at Ensuring Transparency and Reliability of Generative AI Models" project of the Ministry of Education, Culture, Sports, Science and Technology. This study was carried out using the TSUBAME4.0 supercomputer at Institute of Science Tokyo.

\section*{Author Contributions}
Taishi Nakamura initiated the project, designed the method, and carried out the experiments. Takuya Akiba co-designed the experiments and formulated the overall research strategy. Kazuki Fujii implemented the training codebase used for the experiments. Yusuke Oda handled the training of the dense models. Jun Suzuki and Rio Yokota provided guidance and oversight throughout the project. All authors contributed to the writing and approved the final manuscript.

\bibliography{iclr2025_conference}
\bibliographystyle{iclr2025_conference}

\appendix

\clearpage
\section{Experimental Setup Details}
\subsection{FLOPs Calculation}

\begin{table}[t]
\caption{Detailed FLOPs Breakdown for Transformer Models (Single Forward Pass)}
\label{tab:detailed-flops}
\centering
\small
\begin{tabular}{lc}
\toprule
Component & FLOPs \\
\midrule
Embeddings & $2svd_h$ \\
\midrule
Attention (per layer) &  \\
\quad Key and value projections & $4sd_h d_k n_q$ \\
\quad Query projections & $2sd_h d_k n_h$ \\
\quad Key @ Query logits & $2s^2d_k n_h$ \\
\quad Attention matrix computation & $2s^2d_k n_h$ \\
\quad Softmax @ value reductions & $2sd_k n_h d_h $ \\
\midrule
FFN (SwiGLU, per layer) & \\
\quad Dense model & $4sd_h d_f + 2sd_fd_h$ \\
\quad MoE model & $n_e(4sd_h d_f + 2sd_fd_h)$ \\
\midrule
Final Logits & $2sd_hv$ \\
\midrule
Total (Dense) & embeddings $+ n_l(attention + \text{ffn}_{\text{Dense}}) + $ logits \\
Total (MoE) & embeddings $+ n_l(attention + \text{ffn}_{\text{MoE}}) + $ logits \\
\bottomrule
\end{tabular}
\end{table}

Table~\ref{tab:detailed-flops} presents the method for calculating FLOPs (floating point operations) for the forward path in transformer components. The variables used are as follows: $s$ (sequence length), $d_h$ (hidden size), $v$ (vocabulary size), $d_f$ (FFN intermediate size), $n_l$ (number of layers), $n_h$ (number of attention heads), $n_q$ (number of query groups), $d_k$ (attention head dimension), and $n_e$ (number of selected experts per token). For matrix multiplication $A_{m\times k} \times X_{k\times n}$, $2m\times k \times n$ FLOPs are required in the forward pass (the factor of 2 accounts for both multiplication and addition operations). The table displays the main FLOPs contributors for the forward path only. It should be noted that the computational costs for sigmoid and Hadamard product within SwiGLU calculations, MoE gate computations, and RMS Norm calculations are considered negligible and thus omitted from this analysis. While not shown in the table, backward propagation typically requires approximately twice the FLOPs of forward propagation.

% % \label{tab:detailed-flops}にflopsの計算方法を表す. 

% $s$: sequence length 
% $d_h$: hidden size 
% $v$: vocabulary size 
% $d_f$: ffn intermediate size
% $n_l$: number of layers 
% $n_h$: number of attention heads 
% $n_q$: number of query groups  
% $d_k$: attention head dimension
% $n_e$: number of selected experts per token
% である 

% 行列乗算：$A_{m\times k} \times X_{k\times n}$ の行列乗算には $2m\times k \times n$ FLOPsが必要です（乗算と加算を考慮するため2を掛けます）
% 後方伝播は通常, 前方伝播の約2倍のFLOPsを必要とする. 
% 主なflops貢献を　
% 表示する 

% (SwiGLU計算内でのシグモイド, アダマール積, MoEのGateの計算, RMS Normの計算量は無視できるレベルに小さいので無視する. )

% \small
% Variables: \\
% $s$: sequence length (4096),
% $d_h$: hidden size (3072),
% $v$: vocabulary size (99574), \\
% $d_f$: intermediate size (8192),
% $n_l$: number of layers (28), \\
% $n_h$: number of attention heads (24),
% $n_q$: number of query groups (24), \\
% $d_k$: attention head dimension (128),
% $n_e$: number of selected experts per token (2)
% \normalsize
\

\subsection{Implementation and Training environment}\label{appendix:training_environment}

% \Taishi{First Pass}

For our experiments with MoE models and the training of the 1.5B Dense model, we utilized the TSUBAME 4.0 supercomputer at the Global Scientific Information and Computing Center, Institute of Science Tokyo. This environment is equipped with NVIDIA H100 SXM5 94GB GPUs, with each node housing 4 H100 GPUs. Inter-node communication is facilitated by InfiniBand NDR200 interconnects. The training of our largest model, the 8×3.7B model, employed 16 nodes (totaling 64 GPUs).
For the training of the 152M and 3.7B Dense models, we leveraged the high-performance computing nodes (PHY) provided by Sakura Internet. This setup features NVIDIA H100 80GB GPUs, with each node containing 8 H100 GPUs. The network interface is equipped with four 400Gb RoCEv2-compatible NICs and two 25Gb NICs. The training of our largest Dense model (3.7B parameters) utilized a maximum of 32 nodes (totaling 256 GPUs).

For implementation, we used Megatron-LM\footnote{\url{https://github.com/NVIDIA/Megatron-LM}} for Dense model training, and moe-recipes\footnote{\url{https://github.com/rioyokotalab/moe-recipes}, Version 1.0.0} for MoE model training. 
Additionally, Flash Attention 2 \citep{dao2023flashattention2} was utilized to improve computational efficiency and reduce memory usage.
All the training processes were conducted using bfloat16 precision.

% MoEモデルの実験と1.5BのDenseモデルの学習に, 東京工業大学学術国際情報センターのTSUBAME 4.0スーパーコンピューターを利用した. この環境では, NVIDIA H100 SXM5 94GB GPUを使用し, 各ノードに4基のH100 GPUが搭載されている. ノード間はInfiniBand NDR200で相互接続されている. 本研究で使用した最大モデルサイズである8×3.7Bモデルの学習には, 16ノード（計64基のGPU）を使用した. 152Mと3.7BのDenseモデルの学習には, さくらインターネットの高火力コンピューティングノード（PHY）を利用した. この環境では, NVIDIA H100 80GB GPUを使用し, 各ノードに8基のH100 GPUが搭載されている. ネットワークインターフェースには400Gb RoCEv2対応NICが4基と25Gb NICが2基搭載されている. 本研究で使用した最大のDenseモデルサイズ3.7Bの学習には最大32ノード（計256基のGPU）を使用した. 

\subsection{Model configurations}\label{appendix:model_configs_details}

% \begin{table}[t]
% \caption{Model Configuration Details}
% \label{tab:model-details}
% \centering
% \small
% \begin{tabular}{lcccccccc}
% \toprule
% \textbf{Model} & \textbf{Params} & \textbf{Active} & \textbf{Layers} & \textbf{d\textsubscript{model}} & \textbf{d\textsubscript{ff}} & \textbf{Attn} & \textbf{KV} & \textbf{Vocab} \\
%  &  &  & & & & \textbf{Heads} & \textbf{Heads} & \textbf{Size} \\
% \midrule
% Dense 152M & 152 & 152 & 12 & 512 & 2,048 & 8 & 8 & 99,574 \\
% Dense 1.5B & 1,500 & 1,500 & 24 & 2,048 & 7,168 & 16 & 8 & 48,586 \\
% Dense 3.7B & 3,700 & 3,700 & 28 & 3,072 & 8,192 & 24 & 24 & 99,574 \\
% Dense 13B & 13,000 & 13,000 & 40 & 5,120 & 13,824 & 40 & 40 & 99,574 \\
% \midrule
% MoE 8×152M & 420 & 190 & 12 & 512 & 2,048 & 8 & 8 & 99,574 \\
% MoE 8×1.5B & 8,960 & 2,600 & 24 & 2,048 & 7,168 & 16 & 8 & 48,586 \\
% MoE 8×3.7B & 18,600 & 5,900 & 28 & 3,072 & 8,192 & 24 & 24 & 99,574 \\
% \bottomrule
% \end{tabular}
% \end{table}

\begin{table}[t]
\caption{Model Configuration Details}
\label{tab:model-details}
\centering
\small
\begin{tabular}{lccccccc}
\toprule
\textbf{Model} & \textbf{Act Params /} & 
\textbf{Layers} & \textbf{d\textsubscript{model}} & \textbf{d\textsubscript{ff}} & \textbf{Attn} & \textbf{KV} & \textbf{Vocab} \\
 & \textbf{Total Params} & & & & \textbf{Heads} & \textbf{Heads} & \textbf{Size} \\
\midrule
Dense 152M & 152M / 152M & 12 & 512 & 2,048 & 8 & 8 & 99,574 \\
Dense 1.5B & 1.5B / 1.5B & 24 & 2,048 & 7,168 & 16 & 8 & 48,586 \\
Dense 3.7B & 3.7B / 3.7B & 28 & 3,072 & 8,192 & 24 & 24 & 99,574 \\
Dense 13B & 13B / 13B & 40 & 5,120 & 13,824 & 40 & 40 & 99,574 \\
\midrule
MoE 8×152M & 190M / 417M & 12 & 512 & 2,048 & 8 & 8 & 99,574 \\
MoE 8×1.5B & 2.6B / 8.9B & 24 & 2,048 & 7,168 & 16 & 8 & 48,586 \\
MoE 8×3.7B & 5.9B / 18B & 28 & 3,072 & 8,192 & 24 & 24 & 99,574 \\
\bottomrule
\end{tabular}
\end{table}

% \begin{table}[t]
% \caption{実験に使用したモデルの詳細}
% \label{tab:model-details}
% \centering
% \scriptsize
% \begin{tabular}{lrrrrrrrr}
% \toprule
% モデル名 & \multicolumn{2}{c}{パラメータ数} & レイヤー & 隠れ層 & 中間層 & \multicolumn{2}{c}{Attention} & 語彙 \\
% \cmidrule(lr){2-3} \cmidrule(lr){7-8}
%  & 全体 & アクティブ & 数 & サイズ & サイズ & Head & KV Head & サイズ \\
% \midrule
% Dense 152M & 152M & 152M & 12 & 512 & 2,048 & 8 & 8 & 99,574  \\
% Dense 1.5B & 1.5B & 1.5B & 24 & 2,048 & 7,168 & 16 & 8 & 48,586 \\
% Dense 3.7B & 3.7B & 3.7B & 28 & 3,072 & 8,192 & 24 & 24 & 99,574 \\
% Dense 13B & 13B & 13B & 40 & 5,120 & 13,824 & 40 & 40 & 99,574 \\
% \midrule
% MoE 8×152M & 0.42B & 0.19B & 12 & 512 & 2,048 & 8 & 8 & 99,574 \\
% MoE 8×1.5B & 8.96B & 2.6B & 24 & 2,048 & 7,168 & 16 & 8 & 48,586 \\
% MoE 8×3.7B & 18.6B & 5.9B & 28 & 3,072 & 8,192 & 24 & 24 & 99,574 \\
% \bottomrule
% \end{tabular}
% \end{table}

As described in Section~\ref{sec:model-architecture}, we selected the Llama~\citep{touvron2023llamaopenefficientfoundation} and Mixtral~\citep{jiang2024mixtralexperts} architectures for dense and MoE models, respectively, for our experiments. 
Both architectures are based on the Transformer~\citep{NIPS2017_3f5ee243} with several improvements, including RMSNorm~\citep{zhang-sennrich-neurips19}, SwiGLU~\citep{shazeer2020gluvariantsimprovetransformer}, and rotary position embeddings (RoPE)~\citep{su2024roformer}. 
The notable difference in Mixtral (MoE) from Llama (dense) is that all feedforward network (FFN) layers are replaced by sparsely gated MoE layers. Table~\ref{tab:model-details} shows the details of the model configuration.

\subsection{Model training  configurations}\label{appendix:training_configs_details}

As shared settings for training all models, we adopted the following hyperparameters: AdamW optimizer~\citep{loshchilov2019decoupled} with $\beta_1=0.9$, $\beta_2=0.95$, and $\epsilon=10^{-8}$, sequence length of 4096, weight decay of 0.1, and gradient clipping of 1.0. 
The global batch size was set to 1024 for the 1.5B, 3.7B and 13B models, and 512 for the 152M model.

We used cosine decay for learning rate scheduling. For Dense models, the maximum learning rate was set to $3 \times 10^{-4}$, and it decayed to $3 \times 10^{-5}$ over 1,000B tokens for the 1.5B model, and 2,072B tokens for the 152M, 3.7B and 13B models, with the learning rate remaining constant during the final 2000 steps. 
For MoE models, the maximum learning rate was set to $2 \times 10^{-4}$, and it decayed to $2 \times 10^{-5}$ over 500B tokens.
Additionally, to prevent instability in training due to unbalanced routing on the MoE models, a load balancing loss was introduced, with the coefficient unified at 0.02 across all MoE models.

\section{Datasets and evaluation methods}

\subsection{Training dataset details}\label{appendix:dataset-details}

We used the LLM-jp corpus v3\footnote{https://gitlab.llm-jp.nii.ac.jp/datasets/llm-jp-corpus-v3}, an open corpus curated by the LLM-jp working group, for training English and Japanese bilingual language models.
The corpus consists of 1.7T tokens in English, Japanese, and source code with a small amount of Chinese and Korean tokens.
Following the LLM-jp's scheme, some Japanese portion of the corpus is upsampled by 2 to obtain 2.1T training tokens in total.

\begin{table}[t]
\caption{Statistics of the training dataset.}
\label{tab:dataset}
\centering
\small
\begin{tabular}{ll|r}
\toprule
Language & Subset & \#tokens [$\times 10^9$] \\
\midrule
English  & Dolma 1.6 (sampled) \citep{soldaini-etal-2024-dolma} & 945.\hphantom{0} \\
         & Wikipedia                                            &   4.7 \\
\midrule
Japanese & Common Crawl \citep{llmjp2024llmjpcrossorganizationalprojectresearch} & 381.\hphantom{0} \\
         & Kaken                                                &   0.9 \\
         & NDL WARP HTML                                        &   1.3 \\
         & NDL WARP PDF                                         & 207.\hphantom{0} \\
         & Wikipedia                                            &   1.3 \\
\midrule
Chinese  & Wikipedia                                            & 0.8 \\
\midrule
Korean   & Wikipedia                                            & 0.9 \\
\midrule
Code     & The Stack \citep{kocetkov2023the}                    & 114.\hphantom{0} \\
\bottomrule
\end{tabular}
\end{table}

Table \ref{tab:dataset} describes the statistics of the corpus subsets that were used for training data of the Dense and MoE models in our experiments.

Table~\ref{tab:dataset-distribution} details the dataset distribution percentages used for training the different model sizes. The 152M, 3.7B, and 13B models share the same data proportions, while the 1.5B model has slightly different percentages.

\begin{table}[t]
\caption{Dataset Distribution Overview (Percentages)}
\label{tab:dataset-distribution}
\centering
\small
\begin{tabular}{llrr}
\toprule
Language & Subset & 152M/3.7B/13B & 1.5B \\
\midrule
\multirow{2}{*}{English} & Dolma & 45.6\% & 39.7\% \\
& Wikipedia & 0.2\% & 0.5\% \\
\midrule
\multirow{5}{*}{Japanese} & Common Crawl & 36.8\% & 49.5\% 
 \\
& Kaken & 0.1\% & 0.1\% \\
& NDL WARP HTML & 0.1\% & -  \\
& NDL WARP PDF & 11.5\% & -  \\
& Wikipedia & 0.1\% & 0.2\%  \\
\midrule
Chinese & Wikipedia & 0.1\% & - \\
\midrule
Korean & Wikipedia & 0.1\% & -  \\
\midrule
Code & The Stack & 5.5\% & 10.1\%  \\
\midrule
\multicolumn{2}{l}{Total Tokens (B)} & 2,072 & 1,000 \\
\bottomrule
\end{tabular}
\begin{tabular}{@{}p{\linewidth}@{}}

\end{tabular}
\end{table}

% 1.5Bモデルは1Tトークン学習後, 500BトークンをサンプリングしてUpcyclingを行った. 0.15Bと3.7Bモデルは2072.49Bトークン学習途中の1TチェックポイントからUpcyclingを開始し, 表\ref{tab:training-data-others}に示すデータ配分でサンプリングを行った. 

%\subsubsection{Evaluation Datasets}

% \paragraph{JEMHQA}
% JEMHopQA（JEMHQA）は, 元々マルチホップQAタスクとして設計された日本語の自由回答式質問応答データセットです. 私たちのタスク設定では, このデータセットは与えられた質問から直接回答を生成するモデルの能力を評価するために使用されます. JEMHQAの質問は日本語版Wikipediaの情報を使用して作成されており, 多様なトピックとエンティティを確保しています. 
% JEMHQAは, 知識の範囲とその知識を用いた推論による質問応答能力を評価する重要なベンチマークとして機能します. 
% \paragraph{JSQuAD}
% JSQuADは, SQuAD~\citep{rajpurkar2016squad}の日本語版である日本語機械読解データセットです. これは, 文書と質問を読み, 文書内のテキストの一部を回答として抽出する質問応答形式に焦点を当てています. 
% JSQuADは, 日本語版Wikipediaの記事段落を元文書として使用しています. 
% \paragraph{NIILC}
% NIILCは, 日本語の質問応答システム開発を目的として作成されたデータセットで, 百科事典を参照することで回答できる比較的単純な質問が特徴です. これにより, LLMの百科事典的知識を評価するのに有用です. 
% 元のデータセットには質問の種類や回答の手がかり, 回答の記載場所などの追加情報も含まれていますが, 私たちは質問と回答のペアのみを使用しています. 
% \paragraph{XL-Sum}
% XL-Sum日本語版は, BBCニュース記事から収集された大規模要約データセットであるXL-Sumから日本語部分を抽出し, さらに抽象的要約に適したサブセットにフィルタリングして作成されたデータセットです. データセットは, 記事と要約の15-gramオーバーラップ率を計算し, オーバーラップ率の低いペアを選択してフィルタリングされました. 
% 結果として, このデータセットは単に文を抽出するのではなく, 情報を言い換えたり抽象化したりする能力を必要とします. 

\begin{table}[t]

\caption{Evaluation Benchmark Details}
\tabcolsep 3pt%=0.1cm
\label{tab:eval_benchmark_details}
\centering
\tiny
\begin{tabular}{lccccccccccccc}
\toprule
& \makecell[c]{\textbf{JEM} \\ \textbf{HQA}} & \makecell[c]{\textbf{NIILC}} & \makecell[c]{\textbf{JSQ}} & \makecell[c]{\textbf{XL}-\\\textbf{Sum}} & \makecell[c]{\textbf{WMT} \\ \textbf{E$\to$J}} & \makecell[c]{\textbf{WMT} \\ \textbf{J$\to$E}} & \makecell[c]{\textbf{OB} \\ \textbf{QA}} & \makecell[c]{\textbf{TQA}} & \makecell[c]{\textbf{HS}} & \makecell[c]{\textbf{SQ} \\ \textbf{v2}} & \makecell[c]{\textbf{XW}-\\\textbf{EN}} & \makecell[c]{\textbf{BBH}} \\
\midrule
\textbf{Dataset} & JEMHQA & NIILC & JSQuAD & XL-Sum & \multicolumn{2}{c}{WMT20} & OBQA & TriviaQA & HellaSwag & SQuAD2 & XWINO & BBH \\
\textbf{Task} & \multicolumn{2}{c}{QA} & MRC & Summ. & \multicolumn{2}{c}{Trans.} & \multicolumn{2}{c}{QA} & MRC & MRC & Commonsense & Logical \\
& & & & & & & & & & & Reasoning & Reasoning \\
\textbf{Language} & JA & JA & JA & JA & EN$\to$JA & JA$\to$EN & EN & EN & EN & EN & EN & EN \\
\textbf{\# Instances} & 120 & 198 & 4,442 & 766 & 1,000 & 993 & 500 & 17,944 & 10,042 & 11,873 & 2,325 & 6,511 \\
\textbf{Few-shot \#} & 4 & 4 & 4 & 1 & 4 & 4 & 4 & 4 & 4 & 4 & 4 & 3 \\
\textbf{Evaluation Metric} & \multicolumn{3}{c}{Character F1} & ROUGE-2 & \multicolumn{2}{c}{BLEU} & \multicolumn{5}{c}{Accuracy} & CoT Acc. \\
\bottomrule
\end{tabular}
\end{table}

% \begin{table}[t]
% \caption{評価ベンチマークの詳細}
% \label{tab:eval_benchmark_details}
% \centering
% \scriptsize
% \resizebox{\textwidth}{!}{
% \begin{tabular}{|l|c|c|c|c|c|c|c|c|c|c|c|c|}
% \toprule
% 略称 & JEM & NII & JSQ & XL & E$\to$J & J$\to$E & OBQ & TrQ & HeS & SQ2 & XWI & BBH \\
% \midrule
% データセット & JEMHQA & NIILC & JSQuAD & XL-Sum & \multicolumn{2}{c|}{WMT20} & OBQA & TriviaQA & HellaSwag & SQuAD2 & XWINO & BBH \\
% \hline
% タスク & \multicolumn{2}{c|}{QA} & MRC & 要約 & \multicolumn{2}{c|}{翻訳} & \multicolumn{2}{c|}{QA} & \multicolumn{2}{c|}{MRC} & 常識推論 & 論理推論 \\
% \hline
% 言語 & JA & JA & JA & JA & EN$\to$JA & JA$\to$EN & EN & EN & EN & EN & EN & EN \\
% \hline
% 事例数 & 120 & 198 & 4,442 & 766 & 1,000 & 993 & 500 & 17,944 & 10,042 & 11,873 & 2,325 & 6,511 \\
% \hline
% Few-shot数 & 4 & 4 & 4 & 1 & 4 & 4 & 4 & 4 & 4 & 4 & 4 & 3 \\
% \hline
% 評価指標 & \multicolumn{3}{c|}{文字F1} & ROUGE-2 & \multicolumn{2}{c|}{BLEU} & \multicolumn{5}{c|}{Accuracy} & CoT Acc. \\
% \bottomrule
% \end{tabular}
% }
% \end{table}

\subsection{Evaluation Datasets and Methodologies}
\label{appendix:evaluation_details}

Table \ref{tab:eval_benchmark_details} provides detailed information about the evaluations used in our experiments. The evaluation tasks comprise both Japanese and English language assessments. We utilized publicly available evaluation code for our assessments\footnote{\url{https://github.com/swallow-llm/swallow-evaluation}}.

The evaluation tasks are categorized into seven types, such as free-form QA (NIILC~\citep{sekine-etal-2003-niilc}, JEMHQA~\citep{ishi-etal-2023-jemhopqa}), machine reading comprehension (JSQuAD~\citep{kurihara-etal-2022-jglue}, SQuAD2~\citep{DBLP:conf/acl/RajpurkarJL18-squad2}), abstractive summarization (XL-Sum~\citep{hasan-etal-2021-xlsum}), machine translation (WMT'20 En-Ja, Ja-En~\citep{barrault-etal-2020-findings-wmt20}), question answering (OpenBookQA~\citep{mihaylov-etal-2018-openbookqa}, TriviaQA~\citep{joshi-etal-2017-triviaqa}), common sense reasoning (HellaSwag~\citep{zellers-etal-2019-hellaswag}, XWinograd~\citep{DBLP:conf/acl/TikhonovR21-xwino}), and logical reasoning (Big Bench Hard (BBH)~\citep{suzgun-etal-2023-challenging}).
%
%自由記述QA, 機械読解, 機械翻訳, 質問応答, 常識推論には4-shot学習を適用し, 自動要約には1-shot学習, 論理推論には3-shot学習を用いた. なお, 論理推論タスクではChain-of-Thought手法 \citep{wei2022chain} も併せて使用した. 
%各タスクの詳細な評価指標については, 付録\ref{appendix:evaluation_details}に詳述している. 
We used 4-shot prompting for the Free-form QA, machine reading comprehension, machine translation, question answering, and commonsense reasoning tasks, 
1-shot prompting for the abstractive summarization task, 
and 3-shot prompting for the logical reasoning task. 
Moreover, we also applied the Chain-of-Thought method~\citep{wei2022chain} for the logical reasoning task.

\section{Additional Experimental Results and Analysis}

\subsection{Comparison of Gate Initialization Methods}\label{appendix:initialization_details}

We conducted a detailed investigation into the effects of gate initialization on the performance of \NUname{}. An ablation study was performed on five different initialization patterns. Table \ref{tab:gate-initialization-comparison} presents the comparison results of different gate initialization patterns in an 8×1.5B model. Performance was evaluated after training on 50B tokens.

While preliminary experiments had indicated better results with a standard deviation of 0.28, our main experiments revealed that a uniform distribution with a standard deviation of 0.02 achieved the highest average performance across tasks. Based on these results, we adopted a uniform distribution ($\mathcal{U}(-0.0346, 0.0346)$, as the standard method for gate initialization in this study.
It is worth noting that gate initialization may not be a critical factor in model performance, and any initialization that avoids extreme values such as excessively high standard deviations is likely to be sufficient. 

% \begin{table}[t]
% \caption{8×1.5Bモデルのゲート初期化パターン比較（学習トークン数：50B）}
% \label{tab:gate-initialization-comparison}
% \centering
% \small
% \renewcommand{\arraystretch}{1.2}
% \begin{adjustbox}{width=\linewidth}
% \begin{tabular}{l*{6}{r}*{6}{r}r}
% \toprule
% \multirow{2}{*}{\textbf{初期化手法}} & \multicolumn{6}{c}{\textbf{日本語タスク}} & \multicolumn{6}{c}{\textbf{英語タスク}} & \multirow{2}{*}{\textbf{AVG}} \\
% \cmidrule(lr){2-7} \cmidrule(lr){8-13}
% & JEM & NII & JSQ & XL & J$\to$E & E$\to$J & OBQ & TrQ & SQ2 & HeS & XWI & BBH & \\
% \midrule
% GD (0.02, 0) & 46.1 & 37.9 & \textbf{63.6} & 9.2 & 15.4 & 8.1 & 22.4 & \textbf{19.4} & \textbf{41.7} & \textbf{15.6} & 80.0 & 25.9 & 32.1 \\
% GD (0.28, 0) & 50.6 & 38.6 & 54.6 & 9.3 & 15.5 & \textbf{8.3} & 20.6 & 18.4 & 41.1 & 14.3 & 79.8 & 24.7 & 31.3 \\
% Uniform (0.02, 0) & 49.2 & \textbf{38.9} & 61.0 & \textbf{9.7} & \textbf{16.0} & 7.9 & \textbf{23.6} & 18.9 & \textbf{41.7} & 15.5 & \textbf{80.9} & 23.9 & \textbf{32.3} \\
% Uniform (0.28, 0) & 44.6 & 36.3 & 56.3 & 8.6 & 15.5 & 8.1 & 20.6 & 17.7 & 41.0 & 14.6 & 80.0 & \textbf{26.0} & 30.8 \\
% Uniform (0.28, 0.5) & \textbf{51.5} & 36.8 & 55.6 & 9.0 & 15.7 & 7.9 & 21.6 & 18.3 & 41.0 & 15.3 & 80.1 & 25.1 & 31.5 \\
% \bottomrule
% \end{tabular}
% \end{adjustbox}
% \raggedright
% \small
% GD: 正規分布, Uniform: 一様分布, (std, mean): (標準偏差, 平均). 太字は各タスクで最高のスコアを示す. 
% \end{table}

\begin{table}[t]
\caption{Gate Initialization Pattern Comparison for 8×1.5B Models (Training Tokens: 50B)}
\label{tab:gate-initialization-comparison}
\centering
\small
\renewcommand{\arraystretch}{1.03}
\tabcolsep=0.11cm
\begin{adjustbox}{width=\linewidth}
\begin{tabular}{cl*{13}{r}r}
\toprule
& \textbf{Initialization} & \multicolumn{13}{c}{\textbf{Results}} \\
\cmidrule(lr){2-2} \cmidrule(lr){3-15}
\textbf{\#} & \textbf{Distribution} & \textbf{JEM} & \textbf{NII} & \textbf{JSQ} & \textbf{XL} & \textbf{J$\to$E} & \textbf{E$\to$J} & \textbf{OBQ} & \textbf{TrQ} & \textbf{SQ2} & \textbf{HeS} & \textbf{XWI} & \textbf{BBH} & \textbf{AVG} \\
\midrule
1 & $\mathcal{N}(0, 0.02)$ & 46.1 & 37.9 & \textbf{63.6} & 9.2 & 15.4 & 8.1 & 22.4 & \textbf{19.4} & \textbf{41.7} & \textbf{15.6} & 80.0 & 25.9 & 32.1 \\
2 & $\mathcal{N}(0, 0.2887)^*$ & 50.6 & 38.6 & 54.6 & 9.3 & 15.5 & \textbf{8.3} & 20.6 & 18.4 & 41.1 & 14.3 & 79.8 & 24.7 & 31.3 \\
3 & $\mathcal{U}(-0.0346, 0.0346)^\dagger$  & 49.2 & \textbf{38.9} & 61.0 & \textbf{9.7} & \textbf{16.0} & 7.9 & \textbf{23.6} & 18.9 & \textbf{41.7} & 15.5 & \textbf{80.9} & 23.9 & \textbf{32.3} \\
4 & $\mathcal{U}(-0.5, 0.5)$ & 44.6 & 36.3 & 56.3 & 8.6 & 15.5 & 8.1 & 20.6 & 17.7 & 41.0 & 14.6 & 80.0 & \textbf{26.0} & 30.8 \\
5 & $\mathcal{U}(0, 1)$ & \textbf{51.5} & 36.8 & 55.6 & 9.0 & 15.7 & 7.9 & 21.6 & 18.3 & 41.0 & 15.3 & 80.1 & 25.1 & 31.5 \\
\bottomrule
\end{tabular}
\end{adjustbox}
\raggedright
\small
$\mathcal{N}(\mu, \sigma)$: Normal distribution with mean $\mu$ and standard deviation $\sigma$. \\
$\mathcal{U}(a, b)$: Uniform distribution over the interval $[a, b]$. \\
$^*$ $\sigma = \sqrt{1/12} \approx 0.2887$, matches the standard deviation of $\mathcal{U}(0, 1)$. \\
$^\dagger$ Corrected from $\mathcal{U}(-0.0346, 0.0346)$ to match the standard deviation of 0.02.
Bold values indicate the best score for each task.
\end{table}

\subsection{Detailed analysis of expert routing patterns across layers}\label{subsec:detailed_routing_analysis}

For a comprehensive view of routing patterns across all layers, we provide detailed plots of expert routing probabilities for all 24 layers, grouped into early, middle, and late stages. These plots offer a more granular analysis of how routing behaviors evolve throughout the model depth.

Figures~\ref{fig:routing_early} to \ref{fig:routing_late} show the expert routing patterns for all 24 layers of the 8×1.5B MoE models trained with different methods, grouped into early (layers 0-7), middle (layers 8-15), and late (layers 16-23) stages. This comprehensive view allows for a detailed analysis of how routing patterns evolve across the entire model depth.

\begin{figure}[t]
\centering
\includegraphics[width=\textwidth]{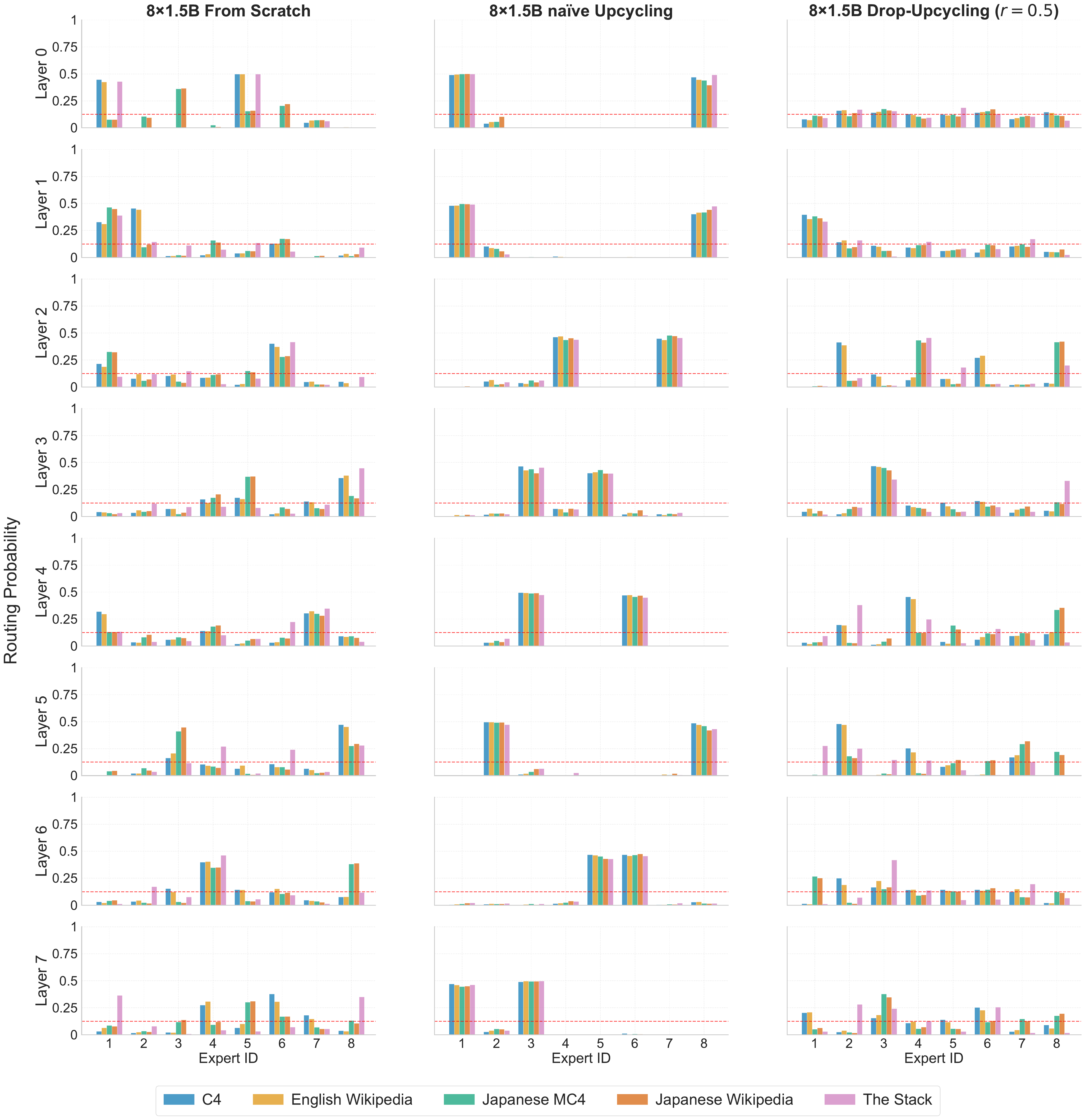}
\caption{Expert routing patterns for early layers (0-7) of the 8×1.5B MoE models.}
\label{fig:routing_early}
\end{figure}

\begin{figure}[t]
\centering
\includegraphics[width=\textwidth]{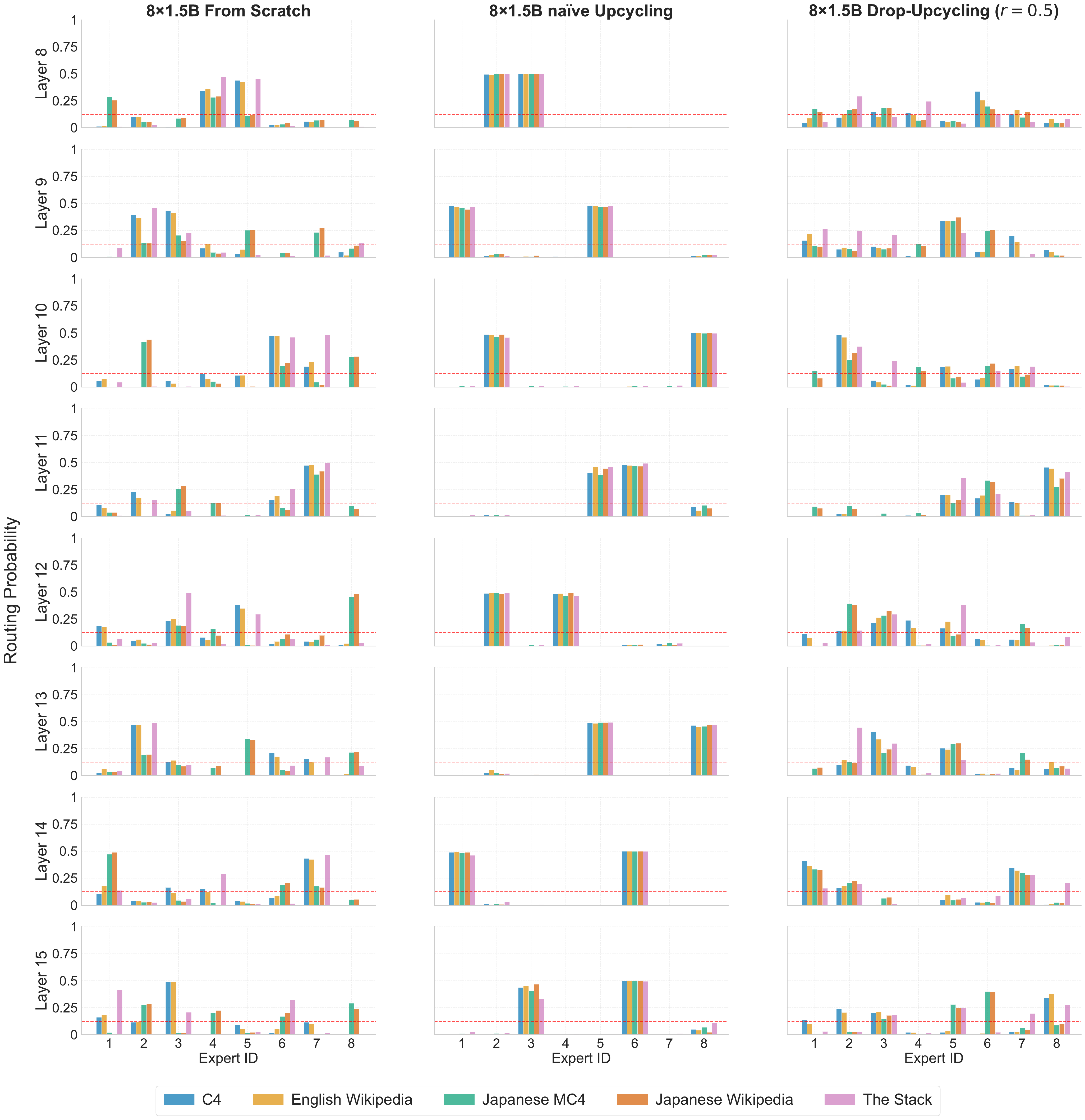}
\caption{Expert routing patterns for middle layers (8-15) of the 8×1.5B MoE models.}
\label{fig:routing_middle}
\end{figure}

\begin{figure}[t]
\centering
\includegraphics[width=\textwidth]{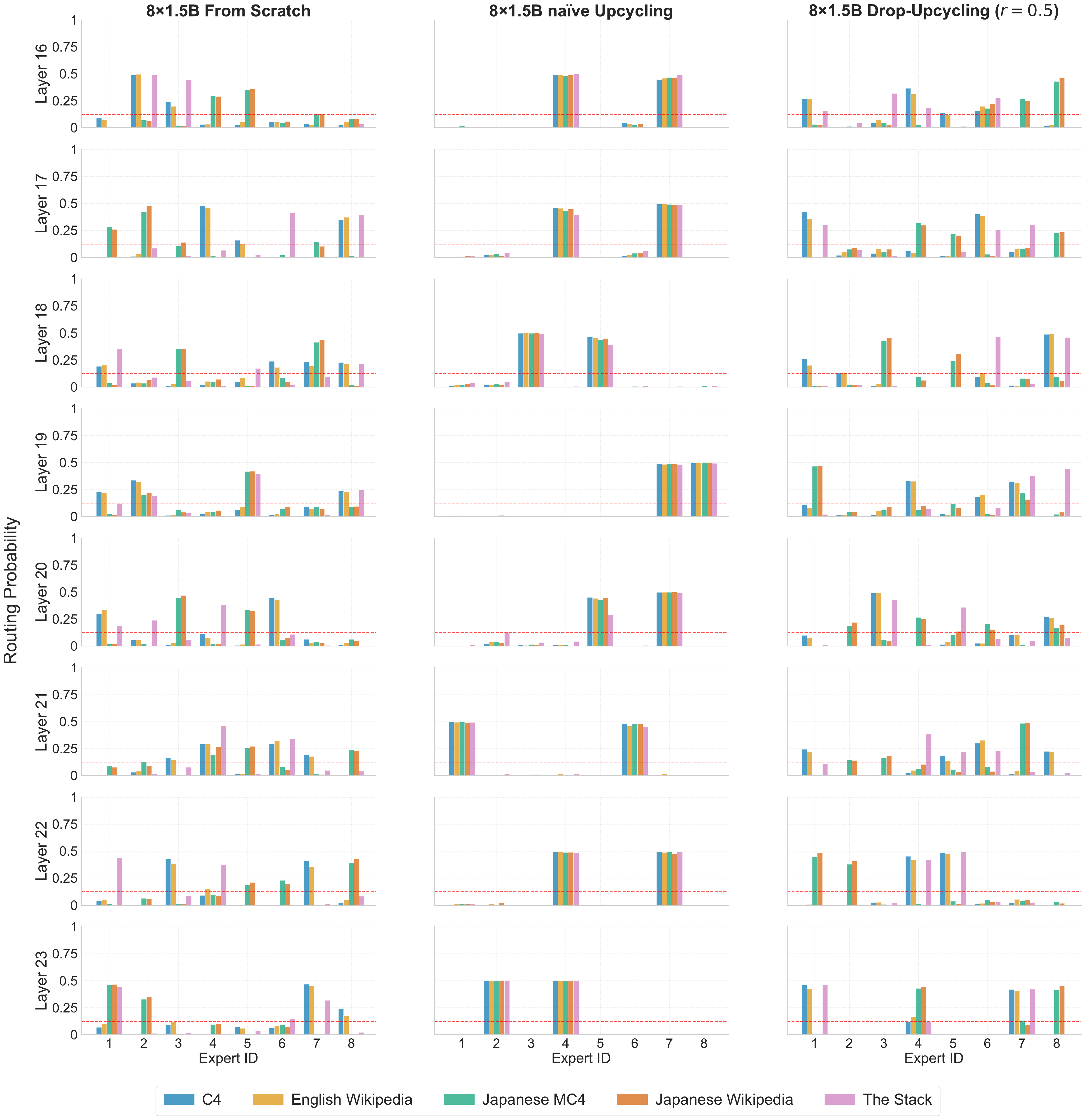}
\caption{Expert routing patterns for late layers (16-23) of the 8×1.5B MoE models.}
\label{fig:routing_late}
\end{figure}

These figures illustrate how the routing patterns evolve throughout the model layers, providing insights into the specialization and behavior of experts at different depths. Notably, the \NUname {} method does not exhibit clear evidence of bias towards specific domains in any layer. In contrast, our proposed method demonstrates domain specialization in multiple layers across the network—from those closest to the input to those near the output—while reusing the parameters of the dense model. This indicates that our approach effectively facilitates expert specialization in several layers without the need to train from scratch, leveraging the pre-trained dense model to achieve efficient domain-specific routing throughout significant portions of the network depth.

\clearpage
\subsection{\diff{Comparing Global vs. Layer-wise Load Balancing}}\label{subsec:detailed_load_balance_comparison}
\diff{In our experiments (Section~\ref{sec:exp}), we applied load balancing loss globally rather than layer-wise. This approach aligns with the implementation in the HuggingFace Transformers library and is widely adopted in the community.
To analyze the effect of global and layer-wise load balancing, we conducted a comparative analysis between global and layer-wise load balancing applications across 40B tokens for different initialization methods (From Scratch, Branch-Train-MiX, \NUname{}, and \methodname{} with r=0.5 and r=1.0) in the 8×1.5B setting.
As shown in Figure~\ref{fig:load_balancing_comparison}, both approaches yield similar training loss trajectories and downstream task performance.
These results suggest that the effectiveness of \methodname{} is not significantly affected by whether load balancing loss is applied globally or layer-wise.}
\begin{figure}[t]
\centering
\includegraphics[width=\textwidth]{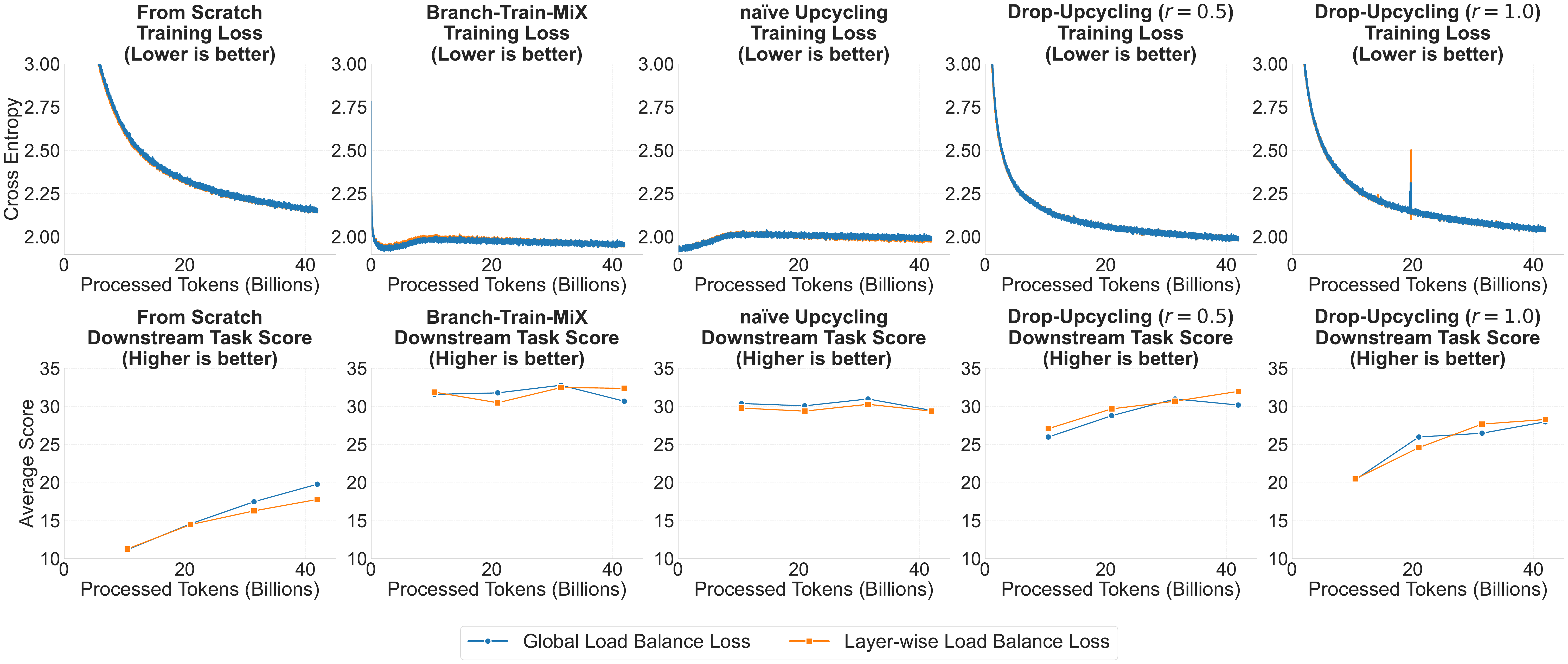}
\caption{\diff{\textbf{Comparison between global and layer-wise load balancing across different initialization methods.} Top: Training loss trajectories over 40B tokens. Bottom: Evaluation metrics measured at iterations corresponding to 10B, 20B, 30B, and 40B tokens. Results show comparable performance between global and layer-wise approaches across all methods.}}
\label{fig:load_balancing_comparison}
\end{figure}

\begin{figure}[t]
\centering
\includegraphics[width=\textwidth]{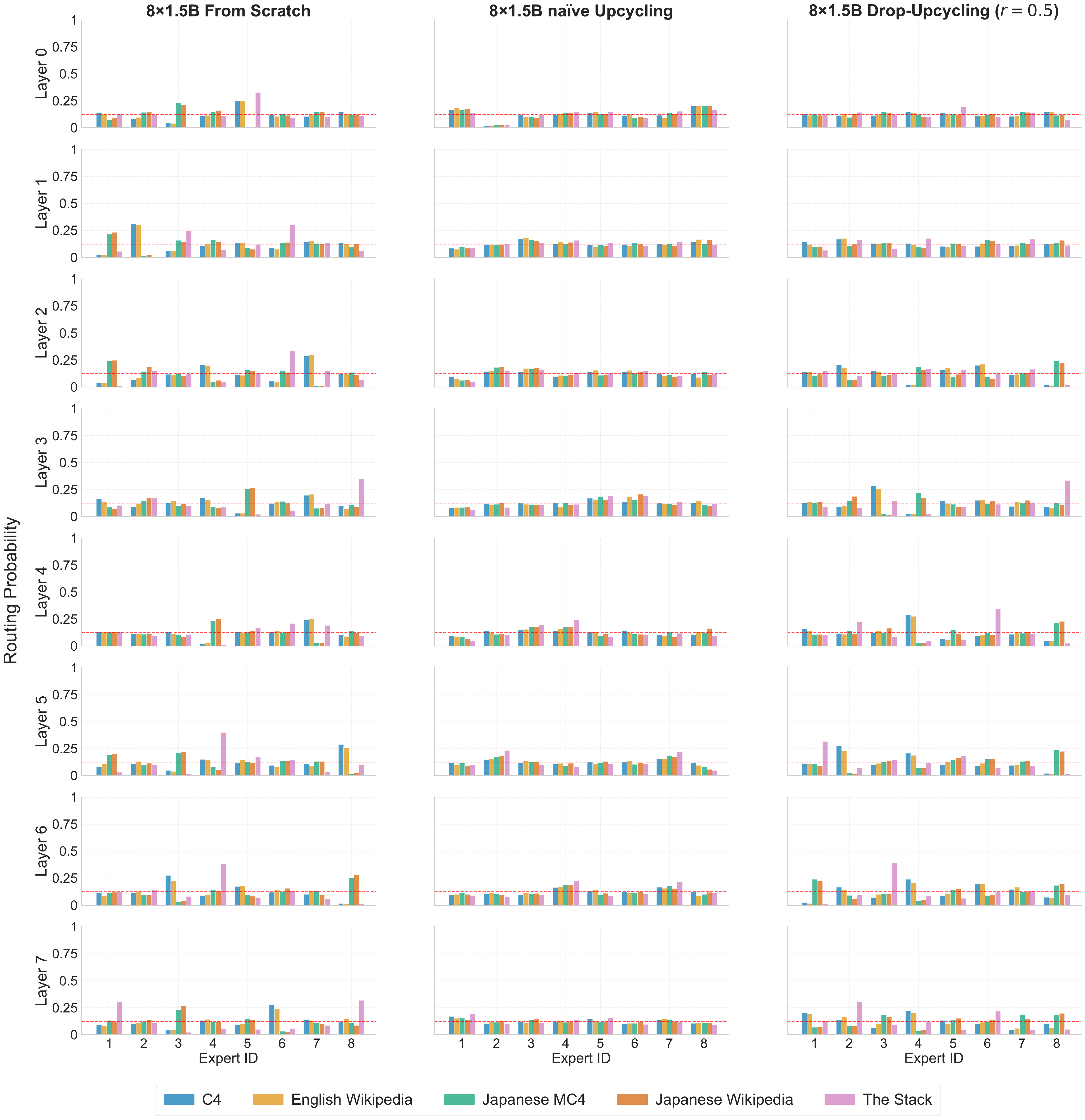}
\caption{\diff{Expert routing patterns for early layers (0-7) under layer-wise load balancing at 40B tokens}}
\label{fig:router_layerwise_early}
\end{figure}

\begin{figure}[t]
\centering
\includegraphics[width=\textwidth]{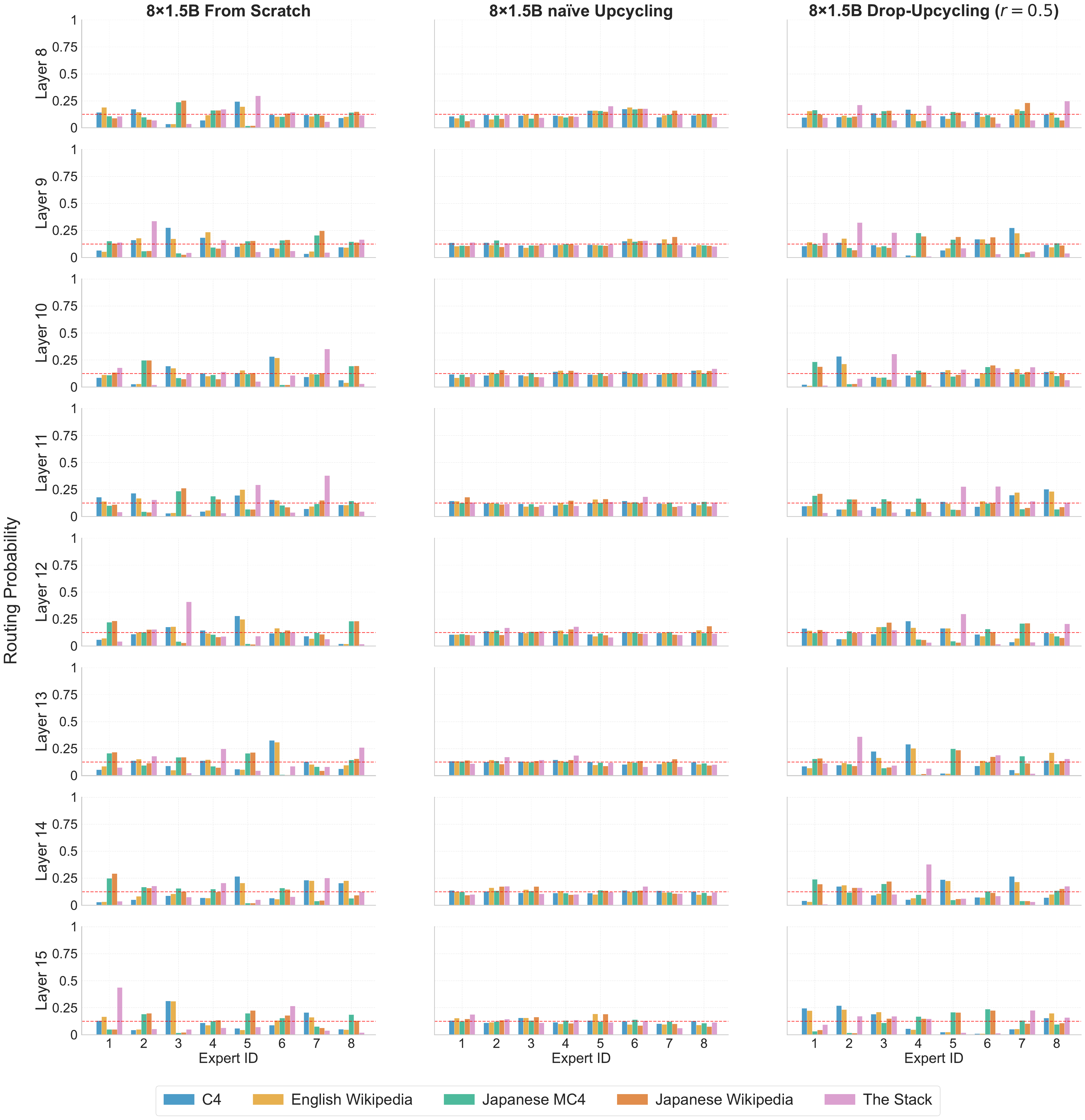}
\caption{\diff{Expert routing patterns for middle layers (8-15) under layer-wise load balancing at 40B tokens}}
\label{fig:router_layerwise_middle}
\end{figure}

\begin{figure}[t]
\centering
\includegraphics[width=\textwidth]{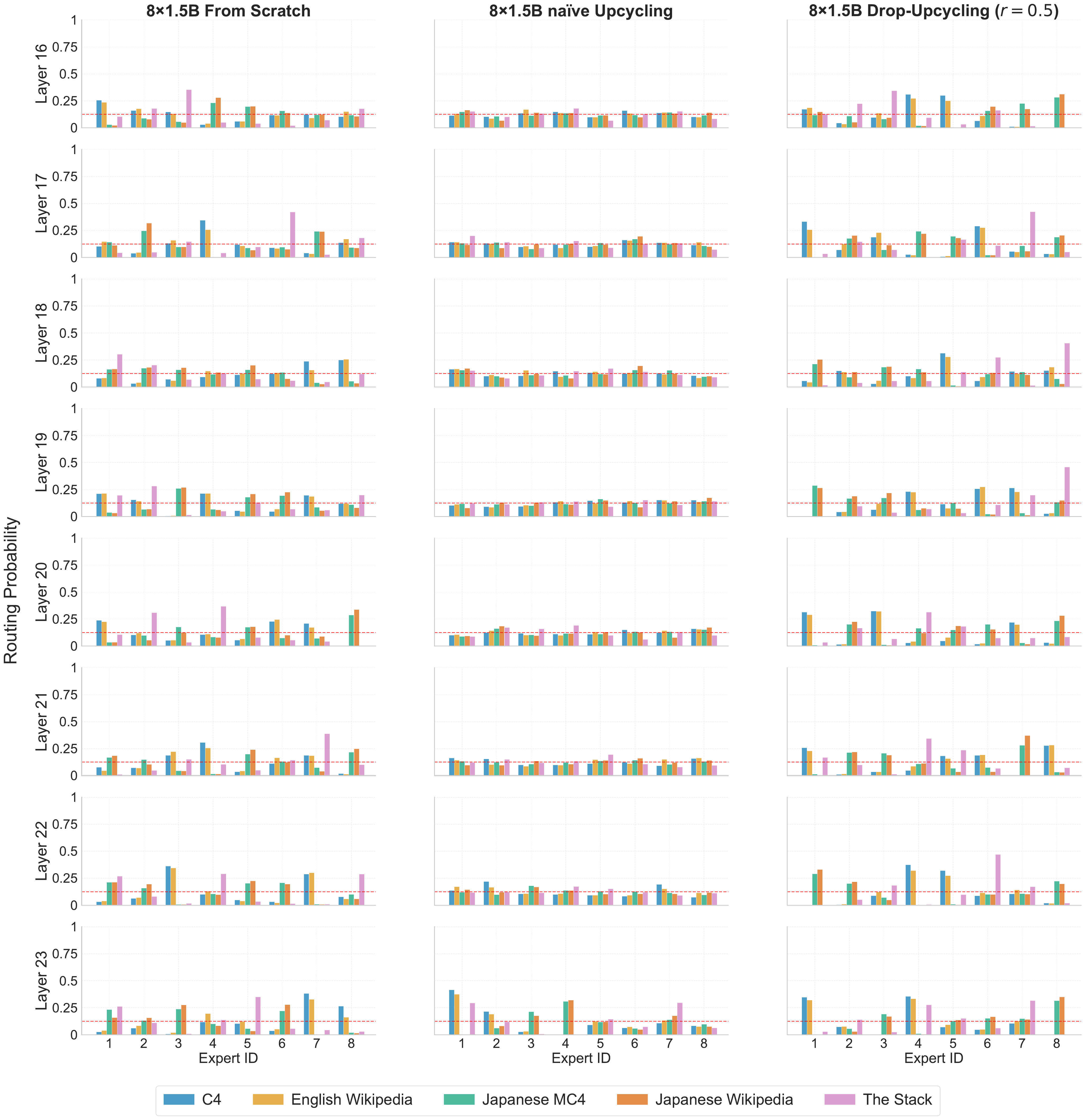}
\caption{\diff{Expert routing patterns for late layers (16-23) under layer-wise load balancing at 40B tokens}}
\label{fig:router_layerwise_late}
\end{figure}

\diff{Figures~\ref{fig:router_layerwise_early} through \ref{fig:router_layerwise_late} show the routing patterns when applying layer-wise load balancing loss at 40B tokens. The results demonstrate that \methodname{} (r=0.5) exhibits domain-specialized routing patterns similar to training from scratch. In contrast, \NUname{} shows nearly uniform routing across all layers except the final layer, which aligns with findings reported in \cite{jiang2024mixtralexperts}. Our proposed \methodname{} method appears to escape the local optima observed in \NUname, which likely contributes to its improved performance.}

\diff{The trade-offs between layer-wise and global load balancing—whether to enforce uniform expert utilization through layer-wise application or to allow potential expert collapse with global application—along with broader questions about MoE architecture design (such as varying expert counts per layer) remain as interesting directions for future research.}

\clearpage
\subsection{\diff{Convergence Catch-Up Analysis}}
\begin{figure}[t]
\centering
\includegraphics[width=\textwidth]{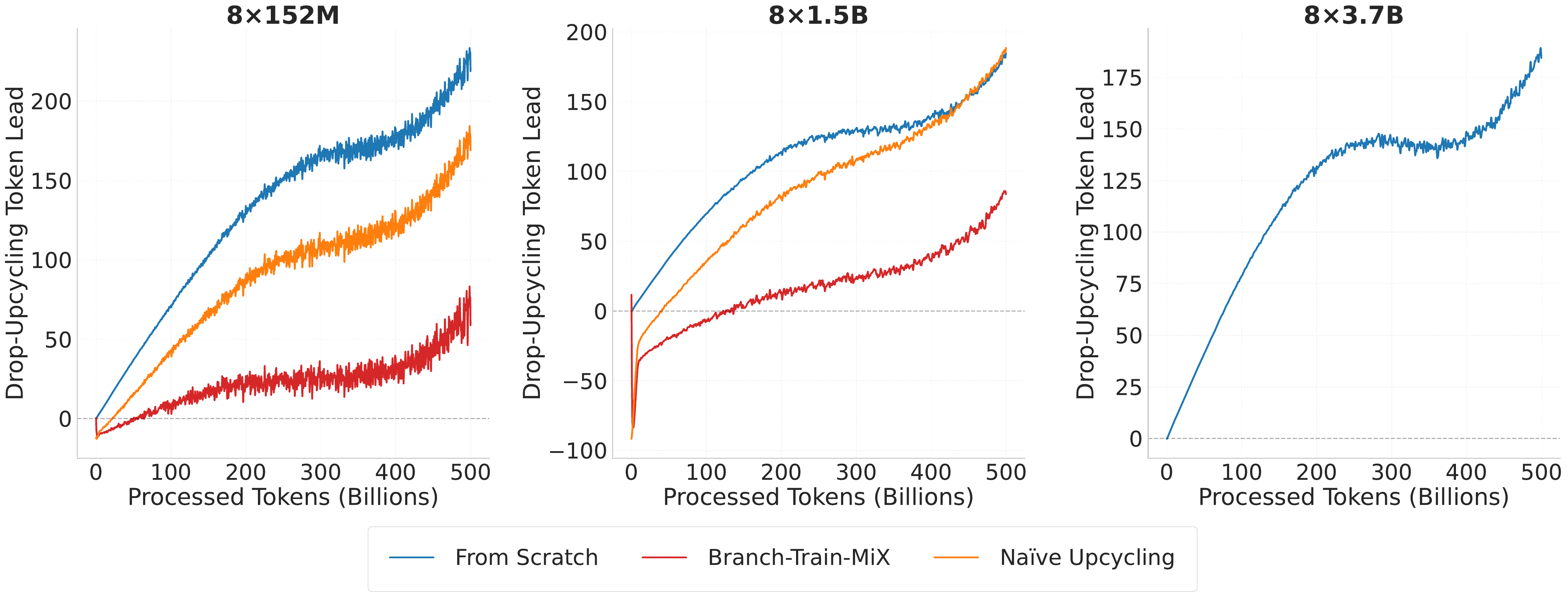}
\vspace{-1em}
\caption{
\diff{\textbf{Convergence catch-up analysis.}
We compare the relative convergence speed of \methodname{} and baseline methods by examining the number of training tokens required to reach the same loss value. The x-axis represents the number of training tokens processed by the baseline method, while the y-axis shows the difference in training tokens needed by \methodname{} to achieve the same loss. Positive values indicate that \methodname{} achieves the loss faster, while negative values suggest the baseline method is ahead.
% The results demonstrate that Drop-Upcycling’s long-term convergence speed is at least comparable to, if not better than, the baseline methods.
}}
\label{fig:convergence_catchup_analysis}
\end{figure}

\diff{To examine the selection of methods based on the training budget and to explore potential extrapolations of long-term trends beyond the scope of our analysis so far, we conduct a brief relative quantitative analysis of the convergence speeds of \methodname{} and baseline methods. In Figure~\ref{fig:convergence_catchup_analysis}, we compare the number of training tokens required to reach the same loss value for \methodname{} and the baseline methods. The plot shows that no significant trend of diminishing advantage for \methodname{} over the baseline methods is observed. This indicates that training from scratch would require an impractically large number of tokens to match \methodname{}, making \methodname{} the better choice in practical scenarios.}

\diff{However, it is important to acknowledge the limitations of this analysis. First, the effect of the learning rate (LR) schedule must be considered. Differences in LR due to different step counts could artificially influence the observed trends in convergence advantage. For example, we hypothesize that the widening advantage of \methodname{} observed late in training (after 400B tokens) may not entirely reflect the contribution of \methodname{} itself but could instead be attributed to the influence of LR scheduling. To eliminate the impact of LR scheduling, conducting all experiments with a constant LR would provide a more valid basis for this comparison.}

\diff{Second, it is worth noting that Branch-Train-Mix utilizes an additional training budget for pretraining individual experts before MoE training. In our setup, for instance, three expert models were pretrained using 100B tokens each, requiring a total of 300B tokens for dense model training before the MoE training phase. As a result, while Branch-Train-Mix appears to show an initial advantage in the plot, this advantage diminishes when accounting for the total training budget. Thus, in terms of overall efficiency, Branch-Train-Mix offers little to no advantage during most of the training process.}

\subsection{\diff{Detailed Derivations of Theoretical Characteristics}}\label{subsec:theoretical}
\diff{Consider the output of MoE layer with parameter re-initialization ratio $r$. Let $\text{FFN}_{\text{retained}_i}(\mathbf{x})$ denote the output from expert $i$'s preserved original parameters (ratio $(1-r)$) and $\text{FFN}_{\text{diverse}_i}(\mathbf{x})$ denote the output from reinitialized parameters (ratio $r$). The exact form of MoE output is:}
\begin{equation}
\diff{\mathbf{y} = \sum_{i=1}^N g(\mathbf{x})_i \cdot (\text{FFN}_{\text{retained}_i}(\mathbf{x}) + \text{FFN}_{\text{diverse}_i}(\mathbf{x}))}
\end{equation}
\diff{where $g(\mathbf{x})_i$ is the gating function defined in~\ref{eq:gating_function}. Note that $g(\mathbf{x})_i = 0$ for experts not among the top-$k$ selected.}

\diff{Let $S_k$ denote the set of indices for the $k$ selected experts. We can rewrite the output as:
\begin{align}
\mathbf{y} &= \sum_{i \in S_k} g(\mathbf{x})_i \cdot (\text{FFN}_{\text{retained}_i}(\mathbf{x}) + \text{FFN}_{\text{diverse}_i}(\mathbf{x})) \nonumber \\
&= \sum_{i \in S_k} g(\mathbf{x})_i \cdot [\text{FFN}_{\text{common}}(\mathbf{x}) + (\text{FFN}_{\text{retained}_i}(\mathbf{x}) - \text{FFN}_{\text{common}}(\mathbf{x})) + \text{FFN}_{\text{diverse}_i}(\mathbf{x})] \nonumber \\
&= \text{FFN}_{\text{common}}(\mathbf{x}) \sum_{i \in S_k} g(\mathbf{x})_i + \sum_{i \in S_k} g(\mathbf{x})_i \cdot [\text{FFN}_{\text{retained}_i}(\mathbf{x}) - \text{FFN}_{\text{common}}(\mathbf{x}) + \text{FFN}_{\text{diverse}_i}(\mathbf{x})] \nonumber \\
&= \text{FFN}_{\text{common}}(\mathbf{x}) + \sum_{i \in S_k} g(\mathbf{x})_i \cdot [\text{FFN}_{\text{retained}_i}(\mathbf{x}) - \text{FFN}_{\text{common}}(\mathbf{x}) + \text{FFN}_{\text{diverse}_i}(\mathbf{x})] \nonumber \\
&= \text{FFN}_{\text{common}}(\mathbf{x}) + \sum_{i=1}^N g(\mathbf{x})_i \cdot [\text{FFN}_{\text{retained}_i}(\mathbf{x}) - \text{FFN}_{\text{common}}(\mathbf{x}) + \text{FFN}_{\text{diverse}_i}(\mathbf{x})]
\end{align}
where the third equality follows from distributing the sum, the fourth equality follows from $\sum_{i \in S_k} g(\mathbf{x})_i = 1$, and the final equality holds because $g(\mathbf{x})_i = 0$ for $i \not\in S_k$. Here $\text{FFN}_{\text{common}}(\mathbf{x})$ represents the output from parameters common to all selected experts.}

\diff{For each expert, a ratio $(1-r)$ of parameters are randomly preserved from the original FFN. When $k$ experts are selected, the probability that a parameter is preserved in all $k$ experts is $(1-r)^k$. Therefore, approximately $(1-r)^k \cdot d_f$ dimensions have common preserved parameters among selected experts, where $d_f$ is the intermediate dimension size. Note that beyond these completely common parameters, there may be partial parameter sharing among subsets of the selected experts due to the random preservation process.}

\diff{To understand the error bound $O(\frac{1}{\sqrt{d_f}})$, consider that for any two experts $i,j$, the number of overlapping parameters follows a binomial distribution $B(d_f, (1-r)^2)$. By the Central Limit Theorem, the deviation from the expected value scales with $\sqrt{d_f}$, leading to a relative error of $O(\frac{1}{\sqrt{d_f}})$ in the parameter overlap estimation.}

\subsection{\diff{Extensions to Fine-grained and Shared Experts}}\label{appendix:extensions}
\diff{We discuss the natural extension of \methodname{} to advanced MoE architectures: fine-grained experts and shared experts proposed in DeepSeekMoE~\citep{dai-etal-2024-deepseekmoe}.
For an original dense FFN with hidden dimension $d_h$ and intermediate size $d_f$, DeepSeekMoE introduces granularity parameter $m$ to split each of $N$ experts into finer segments (each with intermediate size $d_f/m$), where $mk$ experts are selected by top-$mk$ routing, and $k_s$ shared experts process all tokens. The total number of experts becomes $mN$ with $mk$ nonzero gates, which reduces to $mN-k_s$ experts and $mk-k_s$ gates when using shared experts.}

\subsubsection{\diff{Extension to Fine-grained MoE}}
\diff{For simplicity of discussion, we assume $d_f$ is divisible by $m$ for fine-grained MoE (a realistic assumption since $m$ is typically a power of 2 and $d_f$ contains powers of 2 as factors).
The output of the MoE layer is expressed as:}
\begin{equation}
\diff{y = \sum_{i=1}^{mN} g(x)_{(i)} \cdot \text{FFN}_{(i)}(x)}
\end{equation}

%When applying \methodname{} to convert from dense to fine-grained, we consider maximizing diversity in a simple manner.
%From the original dense network, we randomly extract elements from 1 to $d_f/m$ from $d_f$ by rows or columns.
%Note that these elements must not overlap.
%After that, initialize $r$ for each using \methodname{}.

\diff{When applying \methodname{} to convert from a dense FFN layer to a fine-grained MoE layer, we conduct the following steps:}

% \begin{enumerate}
% \item Column-wise Sampling.
% Select an index set $\mathcal{S}$ from the original $d_f$ dimensions for $d_f/m$ dimensions, where $|\mathcal{S}| = \lfloor r \cdot d_f/m \rfloor$.

% \item Statistics Calculation.
% Calculate means and standard deviations $(\mu_\text{up}, \sigma_\text{up})$, $(\mu_\text{gate}, \sigma_\text{gate})$, $(\mu_\text{down}, \sigma_\text{down})$ for the weight matrices corresponding to the selected indices $\mathcal{S}$.

% \item Partial Re-Initialization.
% Initialize each expert's weight matrices according to:
% \begin{equation}
% \widetilde{\mathbf{W}}_{\text{type}} = \mathbf{I}_{\mathcal{S}} \odot \mathbf{R}_{\text{type}} +  (1 - \mathbf{I}_{\mathcal{S}}) \odot \mathbf{W}_{\text{type}}
% \end{equation}
% where $\mathbf{R}_{\text{type}}$ is sampled from $\mathcal{N}(\mu_{\text{type}}, (\sigma_{\text{type}})^2)$.

% \end{enumerate}

\begin{enumerate}
\item \diff{Expert Dimension Sampling.
First, randomly sample $d_f/m$ dimensions from the original FFN intermediate dimension $d_f$ for each expert.}

\item \diff{Column-wise Reinitialization Sampling.
For each expert's sampled $d_f/m$ dimensions, select an index set $\mathcal{S}$ where $|\mathcal{S}| = \lfloor r \cdot d_f/m \rfloor$ dimensions to be reinitialized.}

\item \diff{Statistics Calculation.
Calculate means and standard deviations $(\mu_\text{up}, \sigma_\text{up})$, $(\mu_\text{gate}, \sigma_\text{gate})$, $(\mu_\text{down}, \sigma_\text{down})$ for the weight matrices corresponding to the selected indices $\mathcal{S}$.}

\item \diff{Partial Re-Initialization.
Initialize each expert's weight matrices according to:}
\begin{equation}
\diff{\widetilde{\mathbf{W}}_{\text{type}} = \mathbf{I}_{\mathcal{S}} \odot \mathbf{R}_{\text{type}} +  (1 - \mathbf{I}_{\mathcal{S}}) \odot \mathbf{W}_{\text{type}}}
\end{equation}
\diff{where $\mathbf{R}_{\text{type}}$ is sampled from $\mathcal{N}(\mu_{\text{type}}, (\sigma_{\text{type}})^2)$.}
\end{enumerate}

\diff{Note that the portion reinitialized by our method needs to be scaled down due to the increased number of activated experts in top-$mk$ routing resulting in smaller $g(\mathbf{x})_i$.
While the absolute magnitude information in router outputs might adapt during training,
following \cite{he2024upcyclinglargelanguagemodels},
scaling the weights of $W_{\text{down}}$ and $W_{\text{up}}$ might be beneficial.}

\subsubsection{\diff{Combination with Shared Experts}}
\diff{When using both shared experts and fine-grained experts, the output is:}
\begin{equation}
\diff{y = \sum_{i=1}^{k_s} \text{FFN}_{(i)}(x) + \sum_{i=k_s+1}^{mN} g(x)_{(i-k_s)} \cdot \text{FFN}_{(i)}(x)}
\end{equation}

\diff{Here, shared experts are always active and process dimensions ($d_h$, $d_f/m \cdot k_s$), while fine-grained experts each process $d_f/m$ dimensions.}

% We initialize fine-grained experts using the method described above. We can consider directly copying from the dense FFN or applying \methodname{} to initialize shared experts. We apply weight scaling to both types of experts.

\diff{We initialize fine-grained experts using the method described above. For shared experts, we can either randomly sample $d_f/m \cdot k_s$ dimensions from the dense FFN and directly copy the corresponding weights, or apply \methodname{} to those sampled dimensions. We apply weight scaling to both types of experts.}

\diff{Note that whether shared experts maintain the same functionality as dense remains an open research question, and comparing initialization methods for shared experts is left for future work.}

\subsubsection{\diff{Limitations and Future Directions}}
\diff{While we provide basic extensions of our method to fine-grained and shared expert settings, several important research questions remain unexplored. Our method could serve as a baseline for investigating how knowledge from dense models transfers to these advanced MoE architectures. Specifically, analyzing the transformation process from dense to fine-grained or shared experts could provide valuable insights into how these architectures function and develop specialization. For example, tracking how knowledge is distributed across fine-grained experts during training, or understanding what types of information shared experts learn to capture, could deepen our understanding of these MoE variants. Such analyses could also inform better initialization strategies and architectural choices for future MoE models.}

\end{document}